\documentclass[10pt,twocolumn,letterpaper]{article}

\usepackage[pagenumbers]{cvpr} 

\usepackage{booktabs}
\usepackage{amsmath,amssymb}
\usepackage{enumitem}
\usepackage{tikz}
\usetikzlibrary{arrows.meta,patterns,positioning}
\definecolor{viewc}{RGB}{58,110,200}
\definecolor{holec}{RGB}{230,120,40}
\definecolor{textc}{RGB}{40,150,80}
\definecolor{judgec}{RGB}{125,80,190}
\definecolor{genc}{RGB}{225,175,40}
\usepackage{placeins}

\usepackage{multirow}
\makeatletter
\renewcommand\paragraph{\@startsection{paragraph}{4}{\z@}%
  {0pt}{-1em}{\normalfont\normalsize\bfseries}}
\makeatother
\newcommand{\ourspose}{Ours\,+\,pose\xspace}
\usepackage{pgfplots}
\pgfplotsset{compat=1.18}

\definecolor{cvprblue}{rgb}{0.21,0.49,0.74}
\usepackage[pagebackref,breaklinks,colorlinks,allcolors=cvprblue]{hyperref}

\def\confName{CVPR}
\def\confYear{}

\title{Zero to Four Views: Unified Panorama Generation and\\
Preference RL for Unobserved Regions}

\author{Rui Hong$^{1}$\qquad Xugang Ye$^{2}$\qquad Jana Ko\v{s}eck\'a$^{1}$\qquad Kai Fan$^{2,*}$\\[4pt]
$^{1}$George Mason University\qquad $^{2}$Independent Researcher}

\begin{document}
\maketitle
{\renewcommand{\thefootnote}{\fnsymbol{footnote}}\footnotetext[1]{Corresponding author: \texttt{interfk@gmail.com}}}
\begin{abstract}
Panoramas can serve as shared scene backgrounds for multi-shot video
generation, but most existing panorama generators take text alone or a single
input image. We
present a single model that generates a full equirectangular panorama from a
text caption and $K \in \{0,\dots,4\}$ perspective conditioning views taken by a
four-camera rig at one viewpoint. We fine-tune a
pretrained text-to-image rectified-flow transformer in two stages. In the first stage, each
conditioning view is appended to the token sequence together
with the ID of its camera slot, and the number of views varies
during training, so one set of weights serves every input setting. In the
second stage, we improve the region not covered by any conditioning view. Supervised
training computes its loss on noisy latents of the ground truth and never
scores a panorama the model has generated and decoded, so we propose
Uncovered-Region GRPO (UR-GRPO), a reinforcement-learning recipe that rewards
the candidates a vision--language model prefers in pairwise comparisons on crops
around this region. Because the slot ID alone does not tell the model which way the rig
points, we also offer an optional variant that writes each view's camera
direction into its position encoding, which lets the user decide where each
view appears in the output. With one
conditioning view, our model outperforms the previous state of the art, 360Anything,
trained on the same data and evaluated with the same code, on Laval Indoor and gives mixed results on an
indoor subset of SUN360; the second stage improves the results for every number
of conditioning views.
\end{abstract}

\section{Introduction}
\label{sec:intro}

One way to generate a multi-shot video is to generate several views of each
character~\cite{zero123} and a panorama of the scene, to compose from these and
a text prompt one image for the start of each shot, and to give these images to
a video generation model~\cite{seedance2025}, which produces the video with a cut
at each of them.
Because the panorama shows the room in every direction, each shot can take its
background from whichever direction its camera faces, including reverse angles,
so the room stays consistent across camera angles and shots (supplementary
\cref{fig:storyboard} shows an example); the character views do the same for
the character. Such consistency is hard to obtain when shots are generated on
their own~\cite{videodirectorgpt2024,lct2025,storydiffusion2024}.
This paper addresses generating the panorama.

To make the scene panorama, a user may have only a text description of the
room, or one photo of it, or a few photos taken from the same spot. Existing
panorama generators handle only some of these cases: some take only
text~\cite{panfusion2024,spherediff2025}, others take one
photo~\cite{anything360} or several~\cite{panodiff2023,multiviewpano2025}, and
some take either text or one photo~\cite{cubediff2025,diffusion360,jopano2025}. We
train one model that handles all of them: it takes a text description plus up
to four photos taken from the same spot, facing front, back, up and down, which
we call \emph{conditioning views}, and produces the full $1024{\times}2048$ equirectangular panorama.

We build on FLUX.1-dev~\cite{flux2024}, a $12$B text-to-image model based on a
rectified-flow transformer. The transformer reads one sequence of text tokens and
noisy image tokens; we encode each conditioning view with the same VAE and append
its tokens to this sequence, as in 360Anything~\cite{anything360}. We train it in
two stages.

In the first stage, we fine-tune the whole model with a supervised loss and
give it between zero and four conditioning views per training sample. Each conditioning view is labelled only with its slot in the
four-camera rig, not with the direction it faces. Because the
rig is rotated at random during training, the model cannot tell which way the
scene should face, so where the views land in the output can vary, especially
with one or two views. For users who need each view at a given place, an
optional variant, \ourspose, also gives the model the direction of each view.

In the second stage, we propose Uncovered-Region GRPO (UR-GRPO), which uses
reinforcement learning to improve the \emph{uncovered region}, the part of the
panorama that no conditioning view shows.
Supervised training only learns to denoise ground-truth panoramas and never
looks at the panoramas the model itself generates, so it cannot tell when the
model fills this region poorly. Reward fine-tuning methods for image
generators~\cite{imagereward2023,draft2024,alignprop2023} score the generated
images and back-propagate the score through the VAE decoder into the model;
at our resolution of $1024{\times}2048$ and with a $12$B model, this would cost
far more memory and computation. Reinforcement learning needs only a score for
each generated panorama, and its updates do not pass through the decoder. For
each input, the model generates several panoramas, a vision--language model
compares them in pairs on crops around the uncovered region, and the model is
pushed toward the ones the judge prefers. We use a judge rather than a
comparison with the ground truth because the unseen part of a room can look
many different ways and the ground truth shows only one of them. We also check
how well such a judge agrees with human rankings.

With one conditioning view, our model outperforms the previous state of the art,
360Anything~\cite{anything360}, which we re-implement because its code and
weights are not released, train on the same data and evaluate with the same
code, on Laval Indoor~\cite{lavalindoor} and gives mixed results on an
indoor subset of SUN360~\cite{sun360dataset}; the second stage improves the
results for every number of conditioning views.

\paragraph{Contributions.}
\begin{enumerate}[leftmargin=1.4em,itemsep=1pt,topsep=2pt]
  \item A single model that generates a panorama from a text description and
    zero to four conditioning views taken from the same spot, covering everything from
    text-to-panorama generation to panorama completion.
  \item An analysis showing that when each conditioning view is labelled only with its camera
    slot, the orientation of the output is not determined, and an optional
    variant, \ourspose, that also takes each view's direction so that users can
    decide where each view appears.
  \item UR-GRPO, a reinforcement-learning stage that improves the region no conditioning view covers:
    a vision--language model compares generated panoramas in pairs on crops
    around this region, the update is weighted towards its tokens, and nothing
    is back-propagated through the decoder.
\end{enumerate}

\section{Related Work}
\label{sec:related}

\paragraph{Consistent multi-shot generation.}
VideoDirectorGPT~\cite{videodirectorgpt2024} plans multi-scene videos with a
language model, Long Context Tuning~\cite{lct2025} extends the context of a
video model across shots, and StoryDiffusion~\cite{storydiffusion2024} shares
attention across a sequence of generated images. These methods enforce
consistency inside the generator; we instead produce the scene as a separate
asset, a panorama, that every shot image can draw on. Look
Beyond~\cite{lookbeyond2025} likewise generates a panorama from one image and
uses keyframes taken from it as anchors for a video diffusion model, keeping
novel views consistent along long trajectories.

\paragraph{Text-to-panorama generation.}
Most panorama generators adapt a pretrained text-to-image or text-to-video
model to the equirectangular (ERP) format: PanFusion~\cite{panfusion2024} adds a
panorama branch to Stable Diffusion, Diffusion360~\cite{diffusion360} blends the
left and right edges during denoising, DiT360~\cite{dit360_2025} adds cube-face
supervision, SphereDiff~\cite{spherediff2025} uses a spherical latent,
PanoWan~\cite{panowan2025} extends to video, and SHERPA~\cite{sherpa2026} adapts
a frozen FLUX.1-dev with lightweight adapters. They mainly take text, although
Diffusion360 has a single-image model and DiT360 and PanoWan report
outpainting.

\paragraph{Image-to-panorama generation.}
OmniDreamer~\cite{omnidreamer2022} and PanoDiffusion~\cite{panodiffusion2024}
complete a panorama from a partial one. CubeDiff~\cite{cubediff2025}
generates the six faces of a cube map with a multi-view diffusion model and
uses the input photo as the front face; its perspective-crop evaluation is the
basis of the protocol of~\cite{anything360} that we follow. 360Anything~\cite{anything360} is closest to our
setting: it fine-tunes FLUX without changing the architecture to turn one
perspective image with unknown camera parameters into a full panorama. The
image is encoded by the same VAE and appended to the token sequence, and a
circular encoding at the VAE boundary removes the seam.
PanoDiff~\cite{panodiff2023} accepts one or more unregistered photos, predicts
their angles and completes the resulting partial panorama with a latent
diffusion model; MultiViewPano~\cite{multiviewpano2025} handles one or more
arbitrarily positioned images without training, with a multi-view diffusion
model and pose-aware stitching; and JoPano~\cite{jopano2025} unifies
text-to-panorama and single-view-to-panorama generation in one DiT.
Existing methods target specific input settings---text, one view, several
views registered or projected onto a partial panorama---whereas we train one
model that takes text and any subset of zero to four views as extra tokens; neither CubeDiff,
which fixes the input as the front face, nor 360Anything, which infers the
placement, lets the user choose where the input lands in the output
(\cref{sec:pose}).

\paragraph{Camera pose in multi-view diffusion.}
Outside panoramas, multi-view and novel-view diffusion models usually give the
camera pose of each input image: Zero-1-to-3~\cite{zero123} passes the
relative pose as a vector, CAT3D~\cite{cat3d} attaches a ray direction to every
pixel, and Ray Diffusion~\cite{raydiffusion} predicts the rays themselves.
Panorama models that append views to the token sequence~\cite{anything360}
give no camera pose, and our main model gives only a slot ID. \ourspose\ follows the
multi-view line and gives each view its direction, but encodes it in the
position encoding of the view's tokens rather than as a separate vector or
ray map.

\paragraph{Preference optimization for diffusion models.}
Image generators are tuned toward rewards with policy
gradients (DDPO~\cite{ddpo2023}), direct preference optimization
(Diffusion-DPO~\cite{diffusiondpo2023}) and learned reward models
(ImageReward~\cite{imagereward2023}); group-relative policy optimization (GRPO)~\cite{grpo2024}
replaces a learned value function with the average reward of a group of
samples. Reward back-propagation methods such as
ReFL~\cite{imagereward2023}, DRaFT~\cite{draft2024} and
AlignProp~\cite{alignprop2023} instead require a differentiable score, which a
pairwise judge does not give. Flow-GRPO~\cite{flowgrpo2025} and
DanceGRPO~\cite{dancegrpo2025} adapt GRPO to flow-matching models with a
stochastic sampler, and Pref-GRPO~\cite{prefgrpo2025} rewards win rates from
pairwise preferences, as we do. OneReward~\cite{onereward2025} uses a
vision--language model that compares two images for mask-guided filling, but
back-propagates through a one-step prediction of the clean image. For image
editing, Edit-GRPO~\cite{editgrpo2026} and SpatialFlow-GRPO~\cite{spatialflowgrpo2026}
confine the policy update to regions of the image given by an edit mask or by
region boxes. We follow the policy-gradient line, judge only crops around the
uncovered region with a prompted vision--language model, weight the update
towards the tokens of that region, and down-weight
groups that the judge cannot rank using a paired-comparison model with
ties~\cite{davidson1970}.

\section{Method}
\label{sec:method}

\begin{figure*}[t]
\centering
\resizebox{\linewidth}{!}{%
\begin{tikzpicture}[
  font=\footnotesize, >={Stealth[length=4pt,width=3.5pt]},
  arr/.style={->, line width=0.7pt, draw=black!65},
  lbl/.style={font=\scriptsize, text=black!75, align=center},
  note/.style={font=\tiny, text=black!60, align=center},
  panel/.style={rounded corners=4pt, draw=black!18, fill=black!3, line width=0.6pt},
  ptitle/.style={font=\small\bfseries, anchor=north west, text=black!85},
  view/.style={draw=viewc!90!black, line width=0.6pt, fill=viewc!30},
  tok/.style={draw=white, line width=0.8pt, minimum height=0.42cm, inner sep=0pt, anchor=south west},
  model/.style={rounded corners=3pt, draw=black!55, fill=black!8, line width=0.6pt, align=center},
  pics/erpout/.style={code={%
    \fill[genc!40] (0,0) rectangle (3.0,1.5);
    \fill[viewc!32] (0,1.18) rectangle (3.0,1.5);
    \fill[viewc!32] (0.97,0.34) rectangle (2.03,1.16);
    \fill[viewc!32] (0,0.34) rectangle (0.53,1.16);
    \fill[viewc!32] (2.47,0.34) rectangle (3.0,1.16);
    \draw[black!45, line width=0.5pt] (0,0) rectangle (3.0,1.5);
  }},
]
\fill[panel] (0,0) rectangle (11.2,5.85);
\node[ptitle] at (0.12,5.77) {Stage 1: supervised fine-tuning, $K\in\{0,\dots,4\}$};
\fill[panel] (11.45,0) rectangle (17.6,5.85);
\node[ptitle] at (11.57,5.77) {Stage 2: UR-GRPO, preference RL on $\mathcal{H}$};

\begin{scope}[shift={(0.35,3.3)}]
  \fill[black!10] (0,0) rectangle (3.0,1.5);
  \draw[viewc!85!black, line width=0.7pt]
    (0.03,1.2) rectangle (2.97,1.47)
    (0.97,0.34) rectangle (2.03,1.16)
    (0.03,0.34) rectangle (0.53,1.16)
    (2.47,0.34) rectangle (2.97,1.16);
  \node[note, text=viewc!70!black] at (1.5,1.335) {top};
  \node[note, text=viewc!70!black] at (1.5,0.75) {front};
  \node[note, text=viewc!70!black] at (0.28,0.75) {back};
  \node[note, text=viewc!70!black] at (2.72,0.75) {back};
  \draw[black!45, line width=0.5pt] (0,0) rectangle (3.0,1.5);
  \node[note, text=black!60] at (1.5,0.17) {all of $P$ is the target};
  \node[lbl, anchor=south] at (1.5,1.5) {target: ground truth $P$};
\end{scope}
\draw[arr] (1.85,3.25) -- (1.85,2.98) node[midway, right, lbl] {render views};

\begin{scope}[shift={(0.35,2.55)}]
  \foreach \i/\n/\g in {0/front/1, 1/back/1, 2/top/1, 3/bottom/0} {
    \ifnum\g=1
      \draw[view] (\i*0.78,0) rectangle ++(0.66,0.37);
      \node[note, text=viewc!60!black] at (\i*0.78+0.33,0.185) {\n};
    \else
      \draw[black!40, dashed, line width=0.5pt, fill=white] (\i*0.78,0) rectangle ++(0.66,0.37);
      \node[note, text=black!45] at (\i*0.78+0.33,0.185) {\n};
    \fi
  }
  \node[lbl, anchor=north] at (1.5,-0.04) {conditioning views: any $K$ of 4};
\end{scope}

\node[draw=textc!80!black, fill=textc!14, rounded corners=2pt, line width=0.5pt,
      font=\scriptsize, align=left, text width=2.85cm, inner sep=3pt]
  (cap) at (1.85,0.5) {caption $c$: \emph{``a bright living room with a grey sofa \dots''}};

\begin{scope}[shift={(3.95,2.53)}]
  \node[tok, fill=viewc!75, minimum width=0.68cm] (t1) at (0,0) {\tiny\color{white}$z_{\mathrm{front}}$};
  \node[tok, fill=viewc!75, minimum width=0.68cm] (t2) at (0.68,0) {\tiny\color{white}$z_{\mathrm{back}}$};
  \node[tok, fill=viewc!75, minimum width=0.68cm] (t3) at (1.36,0) {\tiny\color{white}$z_{\mathrm{top}}$};
  \node[tok, pattern=north east lines, pattern color=black!50, minimum width=0.85cm] (tz) at (2.04,0) {};
  \node[font=\tiny, fill=white, inner sep=1pt, rounded corners=1pt] at (tz.center) {$z_t$};
  \node[tok, fill=textc!60, minimum width=0.43cm] (tc) at (2.89,0) {\tiny\color{white}$z_c$};
  \draw[black!40, line width=0.5pt, rounded corners=1pt] (0,0) rectangle (3.32,0.42);
  \node[lbl, anchor=south] at (0.83,0.68) {one token sequence};
  \node[note, anchor=north] at (tz.south) {noisy target};
\end{scope}
\draw[arr] (3.4,2.74) -- (3.92,2.74) node[midway, above, note] {VAE};
\draw[arr] (3.4,4.05) -- (6.415,4.05) node[midway, above, note] {VAE: $z_0$; add noise: $z_t$} -- (6.415,3.2);
\draw[arr] (cap.east) -| (7.055,2.5) node[pos=0.25, above, note] {T5 encoder};

\node[model, minimum width=1.6cm, minimum height=1.3cm] (dit) at (8.35,2.74)
  {\textbf{FLUX.1-dev}\\[-1pt]{\scriptsize transformer}\\[-1pt]{\tiny 12B, fully fine-tuned}};
\draw[arr] (7.3,2.74) -- (dit.west);
\draw[arr] (dit.east) -- (9.5,2.74) node[right, lbl, align=left]
  {flow-matching\\loss on all\\target tokens};

\draw[arr, dashed] (dit.north) |- (11.85,4.6);
\node[lbl, anchor=south] at (9.85,4.62) {Stage-1 weights (frozen)};

\node[model, minimum width=1.75cm, minimum height=0.95cm] (pol) at (12.85,4.6)
  {Stage-1 model\\[-1pt]{\scriptsize + LoRA (rank 32)}};

\begin{scope}[shift={(14.55,4.15)}]
  \foreach \d in {0.24,0.12,0} {
    \pic[scale=0.5] at (\d,\d) {erpout};
    \draw[holec!85!black, line width=0.6pt] (\d+0.19,\d+0.13) rectangle (\d+0.56,\d+0.64)
                                             (\d+0.94,\d+0.13) rectangle (\d+1.31,\d+0.64);
  }
  \node[lbl, anchor=north] at (0.87,-0.03) {$G$ outputs $\hat P_1,\dots,\hat P_G$\\for one input ($K\in\{1,\dots,4\}$)};
\end{scope}
\draw[arr] (pol.east) -- (14.52,4.6);

\node[draw=judgec!80!black, fill=judgec!12, rounded corners=3pt, line width=0.6pt,
      align=center, inner sep=4pt] (jud) at (16.2,1.75)
  {\textbf{VLM judge}\\[-2pt]{\tiny Qwen3-VL-32B}\\[2pt]
   {\scriptsize crops around $\mathcal{H}$}};
\draw[arr] (16.2,3.38) -- (jud.north) node[midway, left, lbl] {pairs A/B,\\both orders};

\begin{scope}[shift={(12.0,1.45)}]
  \foreach \w [count=\i] in {1.55,1.15,0.8,0.35} {
    \fill[genc!40, draw=black!40, line width=0.3pt] (0,{1.25-0.28*\i}) rectangle ++(0.2,0.2);
    \fill[judgec!55] (0.28,{1.27-0.28*\i}) rectangle ++(\w,0.16);
  }
  \node[lbl, anchor=north] at (0.95,0.05) {win rates $\rightarrow$ advantage $\hat A_g$};
\end{scope}
\draw[arr] (jud.west) -- (13.98,1.75 |- jud.west);
\begin{scope}[shift={(13.7,0.08)}]
  \fill[viewc!32, draw=viewc!85!black, line width=0.5pt] (0,0.62) rectangle ++(0.3,0.2);
  \node[lbl, anchor=west] at (0.38,0.72) {covered by the conditioning views};
  \fill[genc!40, draw=genc!85!black, line width=0.5pt] (0,0.31) rectangle ++(0.3,0.2);
  \node[lbl, anchor=west] at (0.38,0.41) {$\mathcal{H}$ generated by the model};
  \draw[holec!85!black, line width=0.6pt] (0,0) rectangle ++(0.3,0.2);
  \node[lbl, anchor=west] at (0.38,0.1) {crop around $\mathcal{H}$, sent to the judge};
\end{scope}
\draw[arr] (12.35,2.75) -- (12.35,4.1) node[pos=0.22, right, lbl, align=left] {policy gradient\\+ KL to Stage 1};
\end{tikzpicture}}
\caption{Overview of training. \textbf{Stage 1} fine-tunes one model for any
number $K$ of conditioning views, with a flow-matching loss on the noisy
ground-truth panorama; each view token carries its camera slot ID (Ours) and,
optionally, its direction (\ourspose). \textbf{Stage 2} trains a LoRA adapter on
the frozen model: a VLM compares $G$ candidates per input in pairs on crops
around the uncovered region $\mathcal{H}$, and the winners are reinforced.}
\label{fig:overview}
\end{figure*}
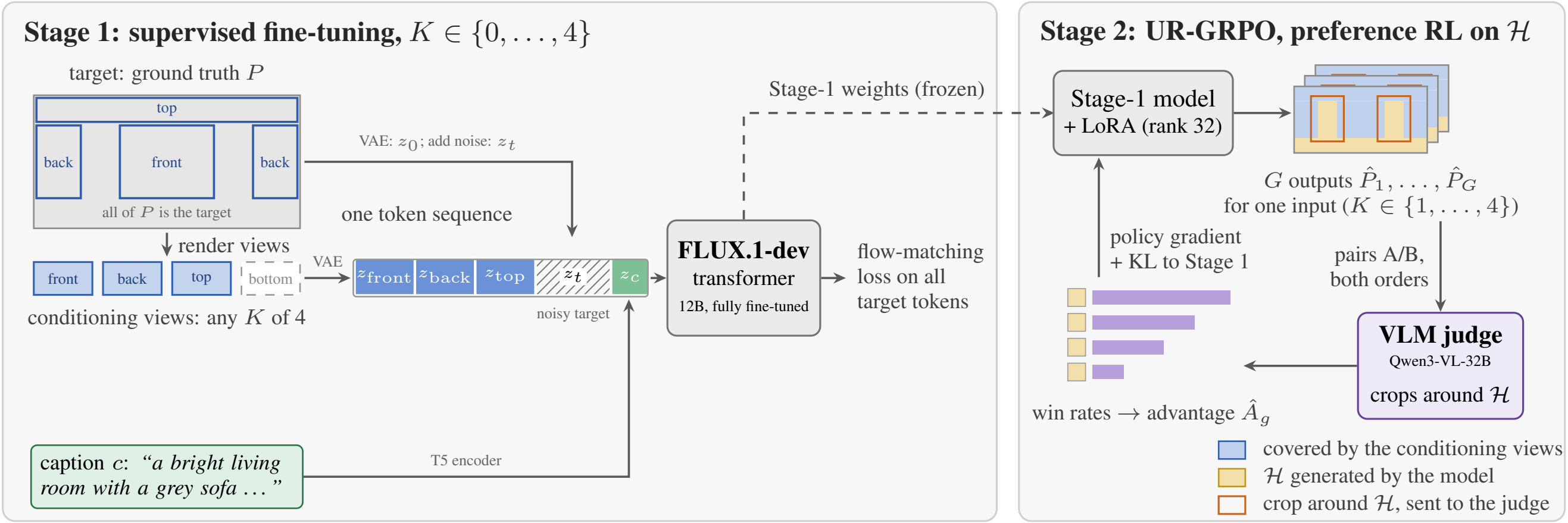

Let $P \in \mathbb{R}^{3 \times H \times W}$ be a gravity-aligned
equirectangular panorama (its horizon is level; $1024{\times}2048$
throughout). The conditioning views are perspective
images rendered from $P$ at its centre by a set of $N{=}4$ cameras looking
front, back, up and down; each view is a $16{:}9$ image about $100^\circ$ to $140^\circ$ wide.
Given a caption $c$ and a subset $\mathcal{K}$ of the views, the task is to
produce the whole panorama:
\begin{equation}
  \hat{P} = f_\theta\!\left(c,\, \{V_i\}_{i \in \mathcal{K}}\right),
  \qquad |\mathcal{K}| = K \in \{0,\dots,4\}.
  \label{eq:task}
\end{equation}
At $K{=}0$ this is text-to-panorama generation; at $K{=}4$, completion of a
mostly observed scene. The uncovered region
$\mathcal{H} = \mathbb{S}^2 \setminus \bigcup_{i \in \mathcal{K}} \Pi_i$ (the
whole sphere at $K{=}0$), where $\Pi_i$ is the part of the sphere seen by view
$i$, splits the output into a part fixed by the inputs and a part that
must be invented. Training has two stages (\cref{fig:overview}): supervised
fine-tuning over all view counts (\cref{sec:stage1}), then UR-GRPO,
reinforcement learning that scores the generated panoramas on the uncovered region
(\cref{sec:rl}).

\subsection{Stage 1: supervised fine-tuning}
\label{sec:stage1}

We fully fine-tune FLUX.1-dev~\cite{flux2024}. The transformer reads one token
sequence, and each token has a three-part rotary position id $(b,i,j)$: $b$
is a slot ID and $(i,j)$ the row and column on the token grid. The views and the
target are turned into tokens by FLUX's VAE and concatenated:
\begin{equation}
  \mathbf{z} \;=\; \big[\, z_{i_1} \cdots z_{i_K} \;\big|\; z_t \,\big],
  \qquad z_i = T(V_i),
  \label{eq:seq}
\end{equation}
where $T$ is FLUX's VAE encoding and tokenization. The noisy target
$z_t = (1{-}t)\,T(P) + t\,\epsilon$ has $b{=}0$, and each view keeps the ID
$b \in \{1,\dots,4\}$ of its slot in the four-camera rig (front $1$, back $2$,
top $3$, bottom $4$). The flow-matching loss is computed on the target tokens
only, and the caption is dropped for a share of the samples so that
classifier-free guidance can be used at test time. A sample contains exactly its
$K$ views and no padding; since each view keeps its slot ID, it is always clear
which views are present, and at $K{=}0$ the caption is the only condition.
During training $K$ is drawn uniformly from $\{0,\dots,4\}$ and the $K$ views
are a random subset of the four, so one model,
which we call \emph{Ours}, covers all view counts.

The camera layout also varies from sample to sample: the camera directions are
slightly perturbed, neighbouring views either all overlap or all leave a gap,
and the whole layout is
rotated at random about the vertical axis and then tilted about a horizontal
axis (\cref{sec:supp-impl} gives the ranges). The uncovered region therefore
changes in size, shape and position from sample to sample; without the rotation
it would always sit at the same image positions (at $K{=}4$, two bands near
azimuth $\pm 90^\circ$).

\paragraph{Reference orientation.} Tilting the layout changes how each view
looks, so the model can read it from the images; a rotation about the vertical
axis only shifts the panorama horizontally, and nothing in the views reveals
it. Because each view carries only its camera slot, the model cannot tell how
far around the panorama the scene should be turned. So in a small share of
training samples we do not rotate the camera layout: the front camera then
always faces the centre of the panorama, and the model learns to put the front
view there by default. \Cref{sec:exp-pose} measures how often it does.

\paragraph{Optional camera directions.}
\label{sec:pose}
Some applications need to decide where each conditioning view appears in the output.
For them, \ourspose\ also gives the model each view's direction: the column
and row ids of the view's tokens are shifted by its longitude and latitude (in
units of the token grid), while its slot ID is kept, so the position ids tell
where the view points and not only which slot it fills. The orientation of the
output then follows these directions (\cref{sec:exp-pose}). The architecture is
unchanged, since position ids are inputs, not weights.

\subsection{Stage 2: uncovered-region GRPO (UR-GRPO)}
\label{sec:rl}

Stage~1 computes its loss on noisy latents of the ground truth and never scores
a panorama the model has generated and decoded. Stage~2 scores decoded samples on the
uncovered region with a vision--language judge and optimizes this score with
group-relative policy optimization~\cite{grpo2024}. Its objective, sampler and win-rate reward follow prior
work~\cite{flowgrpo2025,cps2025,prefgrpo2025}, and weighting the update towards
a region follows recent image-editing methods~\cite{editgrpo2026,spatialflowgrpo2026};
specific to our setting is that the region is the part that no conditioning
view covers and that the reward is judged only on crops around it
(\cref{sec:supp-rlmath}). The score is computed on
decoded images, but the gradient flows only through the log-probabilities of
the latent sampling steps and never through the decoder, and the judge's
pairwise verdicts need not be differentiable. Reward back-propagation
methods~\cite{imagereward2023,draft2024,alignprop2023}, in contrast, pass
gradients from a differentiable score through the decoder and some or all of
the sampling steps, which at $1024{\times}2048$ with a $12$B model would cost
far more memory and computation. We score with a judge rather than with the distance to the
ground truth because the unseen part of the scene can plausibly look many
different ways and the ground truth shows only one of them. The Stage-1 model stays frozen, and a LoRA
adapter of rank $32$ defines the policy $\pi_\theta$ over denoising
trajectories:
\begin{equation}
  \pi_\theta(y \mid x) \;=\; \prod_{t=1}^{T}
  \pi_\theta\!\big(z_{t-1} \mid z_t,\, x\big),
  \label{eq:policy}
\end{equation}
where $x = (c, \{V_i\})$ is the input, the $z_t$ are latent sampler states,
and $y = D(z_0)$ is the decoded sample. Flow-matching sampling is a
deterministic ODE whose steps have no density, so PPO-style policy
gradients~\cite{ppo2017} cannot be applied to it. Our rollouts therefore use a
DDIM-style step~\cite{ddim2021,cps2025} that redraws part of the noise at every
step. Each transition but the last is then Gaussian with known variance,
$\pi_\theta(z_{t-1} \mid z_t, x) =
\mathcal{N}\!\big(\mu_\theta(z_t, t, x),\, \sigma_t^2 I\big)$, so the
log-likelihood of a trajectory in \cref{eq:policy} has a closed form, and the
step preserves the marginal distributions of the training interpolation
(exactly so for exact predictions; proof in \cref{sec:supp-rlmath}).
Each iteration draws inputs with $K \in \{1,\dots,4\}$ from data held out from
Stage~1, generates a group of $G{=}8$ candidates
$\{y_g\}_{g=1}^{G}$ per input, decodes them, and scores them with a pairwise
judgment from a vision--language model (Qwen3-VL-32B) on crops around the
uncovered region. \Cref{sec:supp-stage2} gives the implementation details.

\paragraph{Judgment on the uncovered region.}
\label{sec:judge}
Candidates for the same input are compared in pairs, and a candidate's quality
is its win rate. We compare rather than score because candidates for one input
often differ only in detail: deciding which of two is better is easier and more
consistent for a vision--language model than placing each on an absolute scale,
and the group-relative advantage below needs only the ordering within a group
(we did not test absolute scores). The judge sees perspective crops centred on the
uncovered region, with the region outlined; each crop deliberately includes
some of the covered area around it, so that the judge can tell whether the
generated content fits its surroundings. Each pair is asked in both orders
and both verdicts count, so a preference that depends only on the presentation
order cancels out; the prompt offers a neutrally worded tie option, whose
wording matters (\cref{sec:exp-judge}).

\paragraph{Policy update.} Let $w_g$ be candidate $g$'s win rate, averaged over
the crops of its input, and $\hat z(\cdot)$ the standardization within the
group. The advantage of candidate $g$ is
\begin{equation}
  \hat A_g \;=\; (1 - p_{\mathrm{D}})\, \hat z\big(w_g\big),
  \label{eq:adv}
\end{equation}
where $p_{\mathrm{D}}$ is the $p$-value of a likelihood-ratio test of the
hypothesis that all candidates in the group are equally good, under a
paired-comparison model with ties~\cite{davidson1970}; groups that the judge
cannot rank are thus down-weighted instead of adding noise. The policy is updated by
\begin{equation}
\begin{aligned}
  \nabla_\theta \mathcal{J} =
  \mathbb{E}\Big[\tfrac{1}{G}\textstyle\sum_{g=1}^{G} \hat A_g
  \sum_{t \in \mathcal{T}_g}
  \nabla_\theta \log \pi_\theta\!\big(z^{(g)}_{t-1} \mid z^{(g)}_t, x\big)\Big]&\\
  -\;\beta\, \nabla_\theta\,
  \mathrm{D}_{\mathrm{KL}}\!\big[\pi_\theta \,\Vert\, \pi_{\mathrm{base}}\big]&,
\end{aligned}
  \label{eq:pg}
\end{equation}
with $\mathcal{T}_g$ a random subset of the stochastic sampler steps, drawn anew
for each candidate, and $\pi_{\mathrm{base}}$ the frozen Stage-1 model. In the
log-likelihood, tokens covered by a conditioning view are down-weighted relative to
those in the uncovered region. In our implementation the ratio between the
current policy and the sampling policy stays close to one, so PPO-style
clipping has no effect and \cref{eq:pg} is a plain policy gradient with a
group baseline.

\section{Experimental Setup}
\label{sec:exp}\label{sec:exp-setup}
Full data, training and test details are in \cref{sec:supp-impl}.

\subsection{Training data}
For a fair comparison we train on the same four
panorama sources as 360Anything~\cite{anything360} and
CubeDiff~\cite{cubediff2025}: Structured3D~\cite{structured3d},
Pano360~\cite{spec2021}, PolyHaven~\cite{polyhaven} and Humus~\cite{humus},
$97{,}190$ panoramas in total. The mix differs: Structured3D is $67.4\%$ of our
data and around $90\%$ of the data in~\cite{anything360}. Captions are
generated by Qwen3-VL-32B. The Laval and SUN360 test panoramas are captioned in
the same way from the ground truth, and all models, including the baseline, use
these captions at test time, so a test caption can describe content that the
conditioning views do not show. Stage~1 and Stage~2 use disjoint images ($91\%$
and $9\%$ of the data); Structured3D is split by building, so that Stage~2 does
not judge rooms seen in Stage~1.

\subsection{Test sets}
Following prior work~\cite{cubediff2025,anything360} we
test on \textbf{Laval Indoor}~\cite{lavalindoor} ($2{,}362$ panoramas) and on
\textbf{SUN360}~\cite{sun360dataset}. The SUN360 split used by earlier panorama
papers is no longer available: the official host is offline, our requests to
the dataset authors and to the authors of~\cite{anything360} went unanswered,
and despite our best efforts, including searching mirror sites and public
repositories and contacting the authors of other papers that use SUN360, we
could not find it elsewhere. We therefore use a public set of $518$ indoor SUN360 panoramas
prepared for room-layout estimation~\cite{cfl2020,panocontext2014}, which
differs from the roughly $1{,}000$ indoor and outdoor panoramas used
earlier~\cite{cubediff2025}; these numbers are comparable across the models in
this paper but not with earlier papers. A duplicate search found no Laval panorama in our training
data and $2$ of the $518$ SUN360 panoramas, which we keep in the reported
numbers.

\subsection{Models}
\emph{Ours} and \emph{\ourspose} are trained identically:
Stage~1 fully fine-tunes FLUX.1-dev for $120$k steps, and Stage~2 fine-tunes
the LoRA adapter for $2{,}500$ steps; ``Stage~1+2'' in the tables and figures
denotes the result. The baseline is our re-implementation of
360Anything~\cite{anything360}, also built on FLUX.1-dev and trained for $50$k
steps on the same Stage-1 images with its own recipe; at test time it takes one
$90^\circ$ square view, as in the original method.

\subsection{Inputs at test time}
At $K{=}0$ the model sees only the caption;
at $K{=}1$--$4$ it sees the front view; the front and back views; the front,
back and top views; or all four views. The camera layout is fixed in the
reference orientation, without rotation. At $K{=}1$ our view covers about
$1.43\times$ the solid angle of 360Anything's $90^\circ$ square view, so our
model receives more of the scene. The evaluation crops (below) are taken at
a fixed azimuth range that assumes the view lands at azimuth~$0$; if it does, about
$21\%$ of them overlap the view, so we also report results with crops kept away
from it. The baseline, which has no pose input, places its input at varying
azimuths, and so does Ours when its output is not aligned
(\cref{sec:exp-pose}); for these outputs the crops can include the input
region.

\subsection{Sampling and guidance}
All models, including the baseline, use
the same sampler with $28$ steps. FLUX.1-dev takes a guidance strength as an
input embedding, which we set to $3.5$; rows marked ``+ CFG'' additionally
apply classifier-free guidance (CFG), whose unconditional branch keeps the
conditioning views and drops the caption. We report $K{=}0$ mainly with CFG,
since its outputs look hazy without it, $K{=}1$ with and without CFG, and
$K{=}2$--$4$ without.

\subsection{Metrics}
We follow Table~1 of~\cite{anything360}, using the
implementation of PanFusion~\cite{panfusion2024}. \textbf{Crop FID}, KID and
CLIP-FID compare $10$ perspective crops per panorama, taken at azimuths between
$90^\circ$ and $270^\circ$, with crops of the ground truth. \textbf{FAED}
compares whole panoramas, \textbf{CLIP score} (CS) measures how well the crops
match the caption, and \textbf{DS} measures the discontinuity where the left
and right edges of the panorama meet. All models use the same noise seed for a
given test input, so differences between them are paired.

\section{Results}
\label{sec:results}

\subsection{Comparison with 360Anything at \texorpdfstring{$K{=}1$}{K=1}}
\label{sec:exp-main}

\begin{table*}[t]
\centering\footnotesize
\setlength{\tabcolsep}{3pt}
\begin{tabular}{@{}l cccccc cccccc@{}}
\toprule
& \multicolumn{6}{c}{\textbf{Laval Indoor}} & \multicolumn{6}{c}{\textbf{SUN360 (indoor subset)}} \\
\cmidrule(lr){2-7}\cmidrule(lr){8-13}
Method & crop-FID$\downarrow$ & KID$\downarrow$ & CLIP-FID$\downarrow$ & FAED$\downarrow$ & CS$\uparrow$ & DS & crop-FID$\downarrow$ & KID$\downarrow$ & CLIP-FID$\downarrow$ & FAED$\downarrow$ & CS$\uparrow$ & DS \\
\midrule
\multicolumn{13}{@{}l}{\textit{Guidance embedding only}} \\
360Anything (re-impl.) & $30.66$ & $1.92$ & $8.16$ & $6.25$ & $28.38$ & $0.310$ & $34.06$ & $1.69$ & $10.97$ & $8.07$ & $27.98$ & $0.475$ \\
Ours, Stage 1 & $34.25$ & $2.16$ & $10.85$ & $5.71$ & $27.74$ & $0.565$ & $43.31$ & $2.70$ & $16.12$ & $7.04$ & $27.27$ & $0.670$ \\
Ours, Stage 1+2 & $\mathbf{13.44}$ & $\mathbf{0.64}$ & $\mathbf{3.78}$ & $\mathbf{4.71}$ & $\mathbf{30.17}$ & $0.695$ & $\mathbf{18.45}$ & $\mathbf{0.58}$ & $\mathbf{6.49}$ & $5.30$ & $28.25$ & $0.796$ \\
\ourspose, Stage 1 & $45.41$ & $3.06$ & $13.12$ & $6.12$ & $27.53$ & $0.554$ & $42.91$ & $2.65$ & $15.03$ & $5.78$ & $27.52$ & $0.668$ \\
\ourspose, Stage 1+2 & $19.47$ & $1.07$ & $5.94$ & $5.55$ & $29.78$ & $0.600$ & $20.70$ & $0.79$ & $7.55$ & $\mathbf{4.46}$ & $\mathbf{28.39}$ & $0.748$ \\
\midrule
\multicolumn{13}{@{}l}{\textit{With CFG (scale $3.5$)}} \\
360Anything (re-impl.) & $11.59$ & $0.50$ & $2.66$ & $8.38$ & $30.76$ & $0.365$ & $\mathbf{18.35}$ & $\mathbf{0.69}$ & $\mathbf{5.73}$ & $5.58$ & $28.70$ & $0.529$ \\
Ours, Stage 1 & $14.91$ & $0.68$ & $3.62$ & $7.91$ & $30.28$ & $0.731$ & $23.03$ & $1.06$ & $7.71$ & $4.38$ & $28.21$ & $0.791$ \\
Ours, Stage 1+2 & $\mathbf{10.20}$ & $\mathbf{0.35}$ & $\mathbf{2.43}$ & $6.62$ & $31.51$ & $0.866$ & $21.11$ & $0.87$ & $5.87$ & $\mathbf{3.42}$ & $28.47$ & $0.941$ \\
\ourspose, Stage 1 & $17.76$ & $0.91$ & $3.45$ & $4.83$ & $31.05$ & $0.837$ & $23.17$ & $1.08$ & $6.75$ & $4.38$ & $28.85$ & $0.911$ \\
\ourspose, Stage 1+2 & $14.09$ & $0.64$ & $3.13$ & $\mathbf{4.80}$ & $\mathbf{32.04}$ & $0.921$ & $22.71$ & $1.08$ & $6.55$ & $4.27$ & $\mathbf{28.98}$ & $0.987$ \\
\midrule
Ground truth & & & & & & $0.366$ & & & & & & $0.839$ \\
\bottomrule
\end{tabular}
\caption{Comparison with 360Anything at $K{=}1$ (crops at azimuths $90^\circ$--$270^\circ$). All rows are produced by the same evaluation code on the same test panoramas. 360Anything is our re-implementation, since the original code and weights are not released. Bold: best in each group; DS is not marked because the ground truth itself has a non-zero DS. KID is multiplied by $100$. The SUN360 subset differs from the split used in earlier papers (\cref{sec:exp-setup}).}
\label{tab:main}
\end{table*}

\begin{figure*}[t]
\centering
\includegraphics[width=\linewidth]{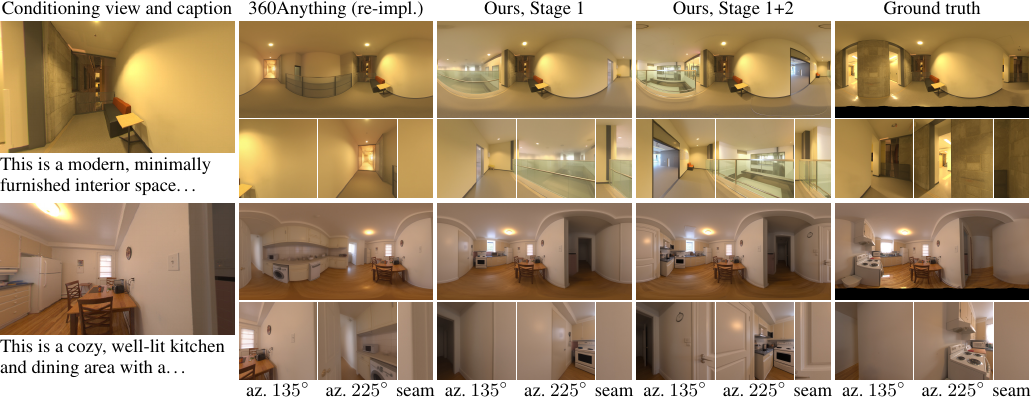}
\caption{Held-out Laval Indoor scenes at $K{=}1$ with CFG (same noise seed for
all models). Left: the conditioning view of our models and the caption
(shortened); the 360Anything re-implementation receives a $90^\circ$ square view
of the same direction. Each other column shows the panorama, two perspective
crops at azimuths $135^\circ$ and $225^\circ$, which face away from the input
when the output is aligned, and a strip around the left/right seam. The scenes
are taken from those where the output of Ours is aligned, so Ours keeps the
input near the panorama centre here; the re-implementation places it at an
arbitrary longitude, so some of its crops show its input region. Selection
rule: \cref{sec:supp-figsel}.}
\label{fig:qual}
\end{figure*}

\Cref{tab:main} compares both of our models with our 360Anything
re-implementation at $K{=}1$, the only setting the baseline supports, and
\cref{fig:qual} shows examples. Both use the same backbone (FLUX.1-dev) and the
same Stage-1 training images, and each follows its own training recipe
(\cref{sec:supp-impl}), so the comparison is between the two complete methods.

\paragraph{Laval.} Before Stage~2 both of our models trail the
re-implementation in crop FID, KID and CLIP-FID. Stage~2 reverses this: Ours is
ahead of the re-implementation on every metric except DS, both without CFG
(crop FID $13.44$ vs.\ $30.66$) and with CFG ($10.20$ vs.\ $11.59$); with CFG,
\ourspose has the best FAED and CLIP score. The crop-FID margin
with CFG ($1.39$) is about as large as the change from re-running the evaluation
with another crop seed ($1.1$), so this is best read as a lead on all five
metrics rather than a large gap.

\paragraph{SUN360.} The results are mixed. Ours after Stage~2 has the best crop
FID, KID and CLIP-FID without CFG, but with CFG the re-implementation has the
best of these three; in both settings our models have the better FAED and
\ourspose the higher CLIP score. CFG helps every model on Laval but hurts our
Stage-2 models on SUN360. We use one guidance setting for all models and test
sets rather than picking the better one per test set.

\paragraph{Is the gain due to the larger conditioning view?} Our $K{=}1$ view
shows more of the scene than the baseline's (\cref{sec:exp-setup}). With crops
kept away from our view and on the held-out Laval scenes (supplementary
\cref{tab:k1crop}),
the ranking on Laval does not change ($11.77$ vs.\ $13.27$ with CFG). This
rules out overlap between the evaluation crops and our view as the source of
the gain, but not the extra scene content that the larger view gives our model.

\paragraph{Published numbers.} Our re-implementation reaches $11.59$ crop FID on Laval
with CFG, against $8.0$ reported for the original model (supplementary
\cref{tab:reported}); differences in
data mix, caption model, inference settings and evaluation preprocessing may
contribute (\cref{sec:supp-impl}). We therefore compare only against the
re-implementation, using one evaluation pipeline for all rows.

\subsection{One model across view counts}
\label{sec:exp-spectrum}

\begin{table*}[t]
\centering\footnotesize
\setlength{\tabcolsep}{3pt}
\begin{tabular}{@{}ll ccccc ccccc@{}}
\toprule
& & \multicolumn{5}{c}{\textbf{Laval Indoor}} & \multicolumn{5}{c}{\textbf{SUN360 (indoor subset)}} \\
\cmidrule(lr){3-7}\cmidrule(lr){8-12}
Input & Model & crop-FID$\downarrow$ & KID$\downarrow$ & CLIP-FID$\downarrow$ & FAED$\downarrow$ & CS$\uparrow$ & crop-FID$\downarrow$ & KID$\downarrow$ & CLIP-FID$\downarrow$ & FAED$\downarrow$ & CS$\uparrow$ \\
\midrule
\multirow{2}{*}{$K{=}0$ + CFG} & Stage 1 & $24.33$ & $1.19$ & $\mathbf{7.49}$ & $16.94$ & $30.19$ & $40.41$ & $3.08$ & $16.10$ & $\mathbf{14.50}$ & $29.12$ \\
 & Stage 1+2 & $\mathbf{16.57}$ & $\mathbf{0.62}$ & $8.15$ & $\mathbf{16.13}$ & $\mathbf{31.39}$ & $\mathbf{35.63}$ & $\mathbf{2.63}$ & $\mathbf{11.54}$ & $14.80$ & $\mathbf{29.61}$ \\
\midrule
\multirow{2}{*}{$K{=}2$} & Stage 1 & $4.88$ & $0.16$ & $1.65$ & $4.20$ & $30.11$ & $9.97$ & $0.32$ & $3.69$ & $2.31$ & $27.71$ \\
 & Stage 1+2 & $\mathbf{3.62}$ & $\mathbf{0.07}$ & $\mathbf{1.03}$ & $\mathbf{3.77}$ & $\mathbf{30.57}$ & $\mathbf{9.07}$ & $\mathbf{0.24}$ & $\mathbf{3.15}$ & $\mathbf{2.01}$ & $\mathbf{27.81}$ \\
\midrule
\multirow{2}{*}{$K{=}3$} & Stage 1 & $3.81$ & $0.11$ & $1.22$ & $3.98$ & $30.38$ & $9.27$ & $0.28$ & $3.25$ & $1.92$ & $27.70$ \\
 & Stage 1+2 & $\mathbf{3.13}$ & $\mathbf{0.06}$ & $\mathbf{0.89}$ & $\mathbf{3.60}$ & $\mathbf{30.68}$ & $\mathbf{8.69}$ & $\mathbf{0.22}$ & $\mathbf{2.86}$ & $\mathbf{1.82}$ & $\mathbf{27.83}$ \\
\midrule
\multirow{2}{*}{$K{=}4$} & Stage 1 & $3.17$ & $0.08$ & $1.05$ & $0.19$ & $30.42$ & $7.91$ & $0.22$ & $2.89$ & $0.85$ & $27.71$ \\
 & Stage 1+2 & $\mathbf{2.80}$ & $\mathbf{0.05}$ & $\mathbf{0.86}$ & $\mathbf{0.13}$ & $\mathbf{30.64}$ & $\mathbf{7.83}$ & $\mathbf{0.21}$ & $\mathbf{2.68}$ & $\mathbf{0.84}$ & $\mathbf{27.79}$ \\
\bottomrule
\end{tabular}
\caption{Ours across view counts, before and after Stage~2. $K{=}0$ uses CFG; $K{=}2$--$4$ do not; $K{=}1$ is in \cref{tab:main}. Bold: the better of the two stages. With two or more views, the inputs already cover most crop directions, so these rows are closer to reconstruction than to generation. All metrics and both model variants are in the supplementary material, with both guidance settings at $K{=}0$ and $K{=}1$; at $K{=}2$--$4$ we did not run CFG.}
\label{tab:spectrum}
\end{table*}

\begin{figure}[t]
\centering
\includegraphics[width=0.78\linewidth]{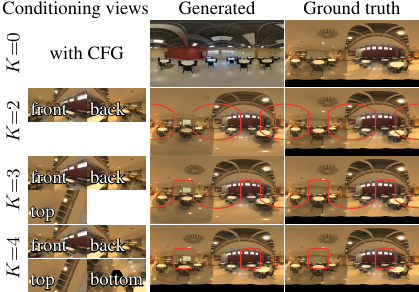}
\caption{Ours (Stage~1+2) on a held-out Laval Indoor scene with zero, two,
three and four conditioning views (same caption and noise seed in all rows).
$K{=}0$ uses CFG; $K{=}2$--$4$ do not and condition on the front and back
views, plus the top view at $K{=}3$ and all four views at $K{=}4$, which
together cover $47\%$, $67\%$ and $87\%$ of the sphere. Red lines mark the
boundary of the area covered by the conditioning views, drawn identically on the
output and on the ground truth. The scene is taken from those whose outputs
are aligned with the ground truth, so that the outline is meaningful
(\cref{sec:supp-figsel}). The black band at the bottom comes from the Laval
images.}
\label{fig:spectrum}
\end{figure}

\Cref{tab:spectrum} reports Ours at every view count, and
\cref{fig:spectrum} shows an example. One set of weights covers every case, from
the caption alone ($K{=}0$) to four views. With two or more views the inputs
already cover most crop directions, so crop FID drops sharply (Laval,
Stage~1+2: $13.44$ at $K{=}1$ to $3.62$ at $K{=}2$) and these settings are
closer to reconstruction than to generation. Stage~2 lowers crop FID at every
view count on both test sets, most where most of the panorama is uncovered
($K{=}0$ and $K{=}1$) and least at $K{=}4$. Although Stage~2 trains only on
$K{=}1$--$4$, it also improves $K{=}0$ (Laval with CFG: $24.33$ to $16.57$).
Results for \ourspose and the remaining metrics are in
\cref{tab:full-laval,tab:full-sun}.

\subsection{Effect of Stage 2}
\label{sec:exp-rl}

\begin{figure}[t]
\centering
\includegraphics[width=0.68\linewidth]{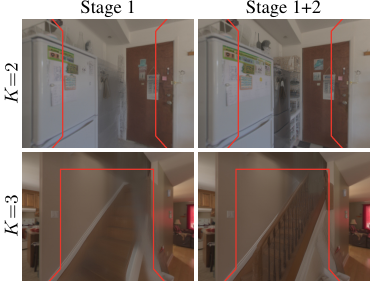}
\caption{The uncovered region of Ours before (left) and after (right) Stage~2,
on held-out Laval scenes at $K{=}2$ and $3$ with the same conditioning views and
noise seed, without CFG ($K{=}1$ is compared in \cref{fig:qual}, and at $K{=}4$
little of the panorama is uncovered). Each row is a perspective crop (horizontal field of view
$85^\circ$) of a side that the conditioning views leave open; the red line
marks the boundary of the uncovered region. We picked these rows because the
Stage-1 result is visibly smooth there (\cref{sec:supp-figsel}); they
illustrate the effect and do not show how often it occurs.}
\label{fig:rl}
\end{figure}

At $K{=}1$ on Laval, Stage~2 lowers the crop FID of Ours from $34.25$ to
$13.44$ without CFG, but CFG alone already closes much of this gap (Stage~1
with CFG: $14.91$). The fair measure of Stage~2 is therefore the comparison with
CFG on both sides: $14.91$ to $10.20$ on Laval and $23.03$ to $21.11$ on
SUN360. After Stage~2, CFG adds less, and on SUN360 it hurts.
\Cref{fig:rl} shows close-ups of the uncovered region before and after
Stage~2; unlike \cref{fig:qual}, it marks the region that no conditioning view
covers.

\subsection{Judge validation}
\label{sec:exp-judge}

\paragraph{Setup.} We check the judge against human rankings. On unseen Laval
scenes, an annotator ranked the eight candidates of each group ($60$ groups in
two batches); we compare only within a group, using which of two candidates the
human prefers. A second pass by the same annotator agrees with the first on
$0.990$ of the pairs; since it was not blind, this is a rough ceiling. Details
are in \cref{sec:supp-judge}.

\begin{table}[t]
\centering\footnotesize
\setlength{\tabcolsep}{2.5pt}
\begin{tabular}{@{}llcc@{}}
\toprule
Judge & tie option & agreement & pairs/group \\
\midrule
Qwen3-VL-32B        & discouraged & $0.622$ {\scriptsize$[.567,.674]$} & --     \\
Qwen3-VL-32B        & neutral     & $0.757$ {\scriptsize$[.701,.805]$} & $11.1$ \\
InternVL3-38B       & discouraged & $0.653$ {\scriptsize$[.585,.716]$} & --     \\
InternVL3-38B       & neutral     & $0.789$ {\scriptsize$[.714,.848]$} & $6.4$  \\
Both judges agree   & neutral     & $\mathbf{0.871}$ {\scriptsize$[.773,.931]$} & $3.0$ \\
Gemma4-12B          & discouraged & $0.393$ {\scriptsize$[.329,.461]$}$^{\ast}$ & -- \\
\midrule
Human self-consistency &          & $0.990$ {\scriptsize$[.978,.996]$} & \\
\bottomrule
\end{tabular}
\caption{Agreement with human rankings ($40$ groups) on pairs the human did not
tie, counting only verdicts that survive swapping the two candidates.
Pairs/group: how many of the $28$ pairs remain. $^{\ast}$Below chance; excluded.}
\label{tab:judge}
\end{table}

\paragraph{Findings (\cref{tab:judge}).} The judge agrees with the human well
above chance ($0.5$): $0.757$ for Qwen3-VL-32B with a neutral tie option, and
$0.871$ when Qwen and InternVL3-38B agree, against a human self-consistency of
$0.990$. Wording the tie option neutrally instead of discouraging ties raises
agreement by $13.5$ points for Qwen and $13.6$ for InternVL; since the gain is
nearly the same for two different models, it is an effect of the prompt rather
than of the model. The wording also changes which position a judge prefers;
Stage~2 asks every pair in both orders, so such a preference cancels in the win
rate. Requiring both judges to agree keeps only $3.0$ of the $28$ pairs per
group, too few to give every candidate a win rate, so training uses
Qwen3-VL-32B alone.

\paragraph{Scope.} The validation differs from Stage~2 (an earlier $K{=}4$
model, a shorter prompt, only order-consistent verdicts;
\cref{sec:supp-judge}). It shows that the judge agrees with human rankings and
that the tie wording matters, but it does not directly measure the accuracy of
the reward or compare pairwise verdicts with absolute scores.

\subsection{Slot ID versus camera direction}
\label{sec:exp-pose}

The two models differ only in whether the position ids contain the camera
directions. At $K{=}1$ Ours has the lower crop FID in every setting of
\cref{tab:main}, while with CFG \ourspose has the best FAED and CLIP score; with
two or more views the two are close (Laval $K{=}2$ crop FID: $3.57$ for
\ourspose, $3.62$ for Ours; \cref{tab:full-laval}). They differ mainly in where the
conditioning views land in the output.

\paragraph{At the test layout.} We measure how far each output is rotated about
the vertical axis relative to the ground truth ($338$ Laval scenes;
\cref{sec:supp-orient}). Ours after Stage~2 is within $10^\circ$ in $57\%$,
$67\%$, $94\%$ and $98\%$ of scenes at $K{=}1$ to $4$, so its orientation is
reliable with three or four views but not with one or two. \ourspose is within
$10^\circ$ in at least $99\%$ of scenes at every $K$. The 360Anything
re-implementation, which has no orientation input, is at chance ($8\%$ against
$5.6\%$ for a uniform offset).

\begin{figure}[t]
\centering
\includegraphics[width=0.8\linewidth]{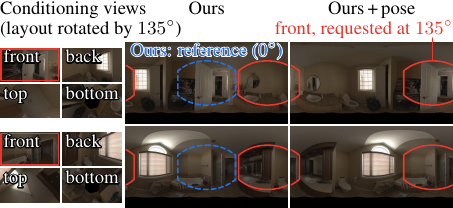}
\caption{Rotated layout ($K{=}4$, no CFG, Stage~1+2, held-out Laval scenes).
The camera layout is rotated by $135^\circ$ about the vertical axis, so the
front view should appear at azimuth $135^\circ$ (red outline), where it also lies
in the ground truth. \ourspose places it there, which puts the back view near
the centre. Ours ignores the rotation and places the front view at its
reference orientation, the panorama centre (blue dashed outline). Scene
selection: \cref{sec:supp-figsel}.}
\label{fig:pose}
\end{figure}

\paragraph{Rotated layouts.} When the whole test layout is rotated about the
vertical axis ($60^\circ$, $135^\circ$ or $240^\circ$; $48$ held-out Laval
scenes; $K{=}1$, $2$ and $4$ without CFG), \ourspose places the front view
within $10^\circ$ of its given direction in every case (\cref{fig:pose}). Ours
has no direction input and moves the views back to its reference orientation:
at $K{=}4$ in $98\%$ of scenes, but at $K{=}1$ and $2$ in only $38$--$60\%$,
with the rest spread over other longitudes. When the layout is also tilted by
$30^\circ$ (one layout), Ours at $K{=}4$ follows the given direction in $71\%$
of scenes, which suggests that views tilted away from the horizon reveal the
orientation; \ourspose stays at $100\%$.

\section{Conclusion and Limitations}
\label{sec:conclusion}

\subsection{Conclusion}
We presented a single model that generates a full panorama from a caption and
zero to four conditioning views. Each view is labelled only with its camera
slot, which lets one set of weights serve every number of views; an optional
variant, \ourspose, also takes each view's direction and places the views where
the user specifies. A second stage, UR-GRPO, rewarded by a
vision--language model that compares candidates in pairs on crops around the
uncovered region, improves the results at every number of views. The model
outperforms the previous state of the art, 360Anything, re-implemented and
evaluated with the same code, on Laval Indoor, and gives mixed results on an
indoor subset of SUN360. Such panoramas can serve as shared backgrounds that
keep the setting consistent across the shots of a generated video.

\subsection{Limitations}
\label{sec:exp-limits}
\textbf{Test data.} Both test sets contain only indoor scenes, and our SUN360
numbers cannot be compared with earlier papers (\cref{sec:exp-setup}); outdoor
scenes remain to be evaluated.
\textbf{Orientation.} With one or two views, Ours does not always place the
views at its reference orientation; when this matters, \ourspose takes the
camera directions as input.
\textbf{Seams.} Our models do not yet treat the seam where the left and right
edges of the panorama meet; a circular latent encoding as
in~\cite{anything360} could be added.
\textbf{Judge scope.} The judge sees crops, not the whole panorama, so it does
not assess global consistency.

\FloatBarrier
{
    \small
    \bibliographystyle{ieeenat_fullname}
    \bibliography{main}
}

\clearpage
\setcounter{page}{1}
\maketitlesupplementary

\section{UR-GRPO: derivation, properties and relation to prior work}
\label{sec:supp-rlmath}
\newtheorem{proposition}{Proposition}
\crefname{proposition}{Prop.}{Props.}
\Crefname{proposition}{Proposition}{Propositions}
\newenvironment{proof}{\par\noindent\emph{Proof.}\ }{\hfill$\square$\par\smallskip}

This section writes out UR-GRPO, the Stage-2 update of \cref{sec:rl}, in full
(\cref{sec:supp-rlmath-def}), states and proves the properties we rely on
(\cref{sec:supp-rlmath-proofs}), and relates each component to prior work
(\cref{sec:supp-rlmath-related}). The name denotes a configuration of existing
components for our setting, not a new reinforcement-learning algorithm: the
objective is that of GRPO for flow
models~\cite{flowgrpo2025,dancegrpo2025}, the sampling step coincides with
coefficients-preserving sampling~\cite{cps2025}, and a win rate from pairwise
comparisons was used as a reward before~\cite{prefgrpo2025}. What is specific
to our setting is where the reward is measured and where the update acts: the
judge sees only crops around the uncovered region (\cref{eq:supp-win}), and the
likelihood is weighted towards the tokens of that region
(\cref{eq:supp-logp}); recent image-editing methods also weight the update by
region~\cite{editgrpo2026,spatialflowgrpo2026}, with the region given by an edit mask
(\cref{sec:supp-rlmath-related}). In addition, groups that the judge cannot rank are
down-weighted by a test (\cref{eq:supp-adv}). \Cref{sec:supp-stage2} lists the remaining
implementation details.

\subsection{The policy and its update}
\label{sec:supp-rlmath-def}

\paragraph{States.} The backbone is a flow-matching model. With $z_0$ the clean
target latent and $\epsilon \sim \mathcal{N}(0, I)$, the training interpolation
is
\begin{equation}
  z_t = (1-t)\, z_0 + t\, \epsilon, \qquad t \in [0, 1],
  \label{eq:supp-interp}
\end{equation}
so $t{=}1$ is noise, and the network predicts the velocity
$v_\theta(z_t, t, x) \approx \epsilon - z_0$ for the input $x$. A latent has
$n = 8192$ target tokens of $d = 64$ channels. Here we index sampler states by
their noise level instead of the step counter of \cref{eq:policy}: the
$T = 28$ steps of \cref{eq:policy} go through the levels $t_k = 1 - k/T$,
$k = 0, \dots, T$, from noise ($t_0 = 1$) to the final latent ($t_T = 0$), and a
step is written as going from a level $t$ to the next level $s < t$. The
variance written $\sigma_t^2$ in \cref{sec:rl} is $\sigma_s^2$ below, the
variance of the step that arrives at level $s$.

\paragraph{One sampling step.} A step from level $t$ to $s < t$ first splits
the state into estimates of its clean and noise parts,
\begin{equation}
  \hat z_0 = z_t - t\, v_\theta, \qquad
  \hat\epsilon = z_t + (1-t)\, v_\theta ,
  \label{eq:supp-split}
\end{equation}
and then recombines them with part of the noise replaced by fresh noise
$\xi \sim \mathcal{N}(0, I)$:
\begin{equation}
\begin{aligned}
  \mu_\theta(z_t) &= (1-s)\, \hat z_0 + s \sqrt{1-\nu^2}\; \hat\epsilon, \\
  z_s &= \mu_\theta(z_t) + \sigma_s\, \xi, \qquad \sigma_s = s\, \nu ,
\end{aligned}
  \label{eq:supp-step}
\end{equation}
with $\nu \in [0, 1]$ the fraction of the noise that is redrawn. Since
$s\sqrt{1-\nu^2} = (s^2 - \sigma_s^2)^{1/2}$, this is the DDIM
update~\cite{ddim2021} with the coefficients $(1-s, s)$ of
\cref{eq:supp-interp} in place of those of a diffusion schedule. Stage-2
rollouts use $\nu = 0.5$ on the uniform levels $t_k$; at test time the same
sampler runs with $\nu = 0.9$ on levels shifted as in
\cref{sec:supp-impl}. The transition is therefore
$\pi_\theta(z_s \mid z_t, x) = \mathcal{N}(\mu_\theta(z_t), \sigma_s^2 I)$. The
last step has $s = 0$, hence $\sigma_s = 0$: it is deterministic and is not part
of the policy, so the product in \cref{eq:policy} has $27$ Gaussian factors
followed by one deterministic map. The candidates of a group
start from independent initial noise.

\paragraph{Why the sampler must be stochastic.} A policy gradient needs the
probability of each sampled transition. The ODE sampler ($\nu = 0$) maps the
initial noise to the output deterministically: its transitions have no density,
and the only randomness, the initial noise, does not depend on $\theta$. The
alternatives are to differentiate the reward through the sampling chain, which
needs a differentiable reward and is costly at this model size
(\cref{sec:rl}), or to compute the likelihood of the ODE output, which needs
the divergence of the velocity along the whole chain. With fresh noise at each
step, every transition is a Gaussian whose log-density is a quadratic form in
one network output, and a gradient step touches one sampling step at a time.

\paragraph{Log-likelihood.} For a stochastic step we use
\begin{equation}
  \ell_\theta(z_s \mid z_t) = - \frac{1}{2 \sigma_s^2}\,
  \frac{\sum_{i=1}^{n} \omega_i\, \tfrac{1}{d} \lVert z_s^{i} - \mu_\theta^{i}(z_t) \rVert^2}
       {\sum_{i=1}^{n} \omega_i},
  \label{eq:supp-logp}
\end{equation}
where $i$ runs over the target tokens and $\omega_i = 0.1 + 0.9\,(1 - c_i)$,
with $c_i \in [0, 1]$ the fraction of token $i$ that a conditioning view covers:
tokens in the uncovered region have weight $1$ and fully covered tokens $0.1$.
The normalizing constant of the Gaussian is omitted because it cancels in the
ratio below.

\paragraph{Reward.} For a crop $c$ and an ordered pair $(g, h)$, in which $g$ is
shown first, let $S_c(g, h) \in \{1, \tfrac12, 0\}$ be the judge's verdict for the
first image (win; tie or unparseable answer; loss). Each pair is asked in both
orders, and the win rate of candidate $g$ is
\begin{equation}
  w_g = \frac{1}{C} \sum_{c=1}^{C} \frac{1}{G-1} \sum_{h \neq g}
        \tfrac12 \Big[ S_c(g, h) + 1 - S_c(h, g) \Big],
  \label{eq:supp-win}
\end{equation}
with $C = 2$ crops and $G = 8$ candidates.

\paragraph{Advantage.} With $\bar w$ and $\mathrm{sd}(w)$ the mean and
(population) standard deviation of the win rates in the group,
\begin{equation}
  \hat A_g = (1 - p_{\mathrm{D}})\; \kappa_G\, \frac{w_g - \bar w}{\mathrm{sd}(w)},
  \quad
  \kappa_G = \Big( \sqrt{\tfrac{G}{G-1}} + 0.05 \Big)^{-1},
  \label{eq:supp-adv}
\end{equation}
and $\hat A_g = 0$ for all $g$ if the win rates are all equal. This is
\cref{eq:adv} with $\hat z(w_g) = \kappa_G (w_g - \bar w)/\mathrm{sd}(w)$;
$\kappa_G \approx 0.89$ is a constant of our implementation that only rescales
the learning rate. The weight $p_{\mathrm{D}}$ comes from the paired-comparison
model with ties of Davidson~\cite{davidson1970}, in which candidates have
strengths $\pi_g > 0$ and a tie parameter $\vartheta > 0$:
\begin{equation}
  \Pr[g \succ h] = \frac{\pi_g}{D_{gh}}, \quad
  \Pr[\text{tie}] = \frac{\vartheta \sqrt{\pi_g \pi_h}}{D_{gh}},
  \label{eq:supp-davidson}
\end{equation}
with $D_{gh} = \pi_g + \pi_h + \vartheta \sqrt{\pi_g \pi_h}$. We fit the model
to the $G(G-1)$ verdicts of a crop by maximum likelihood with a small ridge
penalty on the log-strengths (a candidate that wins every comparison would
otherwise have an infinite estimate), and compare its log-likelihood with that
of equal strengths at the same $\vartheta$. Twice the difference is referred to
a $\chi^2$ distribution with $G - 1$ degrees of freedom, and $p_{\mathrm{D}}$ is
the smallest $p$-value over the crops ($p_{\mathrm{D}} = 1$ if no verdict is
decisive).

\paragraph{Objective.} For candidate $g$ and a stochastic step $(t, s)$ of its
trajectory, let $\rho_\theta = \exp\!\big( \ell_\theta(z_s \mid z_t) -
\ell_{\theta_{\mathrm{old}}}(z_s \mid z_t) \big)$, where $\theta_{\mathrm{old}}$
are the weights that generated the trajectory (the exponent is clamped to
$[-5, 5]$ for numerical safety, which never took effect). The loss of this step is
\begin{equation}
\begin{aligned}
  \mathcal{L} = {} & - \min\!\big( \rho_\theta \hat A_g,\;
      \mathrm{clip}(\rho_\theta, 1 - \varepsilon, 1 + \varepsilon)\, \hat A_g \big) \\
    & + \beta\, \frac{1}{2 \sigma_s^2}\, \frac{1}{n d}\,
      \big\lVert \mu_\theta(z_t) - \mu_{\mathrm{base}}(z_t) \big\rVert^2 ,
\end{aligned}
  \label{eq:supp-loss}
\end{equation}
with $\varepsilon = 0.2$, $\beta = 0.01$, and $\mu_{\mathrm{base}}$ the mean of
\cref{eq:supp-step} computed by the frozen Stage-1 model (the adapter switched
off) at the same state. The first term is the clipped surrogate of
PPO~\cite{ppo2017} with the group-relative advantage of
GRPO~\cite{grpo2024}; the second is the KL divergence of
\cref{prop:kl}, averaged over dimensions and not weighted by $\omega$. The loss
is averaged over the $7$ of $27$ stochastic steps drawn for each candidate and
over the $64$ candidates of an iteration ($8$ inputs, $8$ candidates each),
$448$ transitions in all, and $4$ such gradient steps are taken per batch of
rollouts with fresh draws of the steps (AdamW, learning rate $10^{-4}$, gradient
clipped to norm $1$). Only the mean $\mu_\theta(z_t)$ is recomputed with the
current weights; $z_t$, $z_s$ and $\ell_{\theta_{\mathrm{old}}}$ are those stored
during the rollout, and all selected steps of a candidate share its advantage. \Cref{eq:pg} is the negative gradient of this loss at $\rho_\theta = 1$, up
to constant factors: $\ell_\theta$ stands for the log-likelihood written
$\log \pi_\theta$ there, from which it differs by the average over dimensions
and the token weights (\cref{prop:pg}), the KL term is likewise averaged over
dimensions, and the sum over the selected steps is an average.

\subsection{Properties}
\label{sec:supp-rlmath-proofs}

\begin{proposition}\textup{(Coefficient preservation.)}
\label{prop:cp}
Let $z_t = (1-t) z_0 + t\, \epsilon$ with $\epsilon \sim \mathcal{N}(0, I)$
independent of $z_0$, and suppose the estimates in \cref{eq:supp-split} are
exact, $\hat z_0 = z_0$ and $\hat\epsilon = \epsilon$. Then the state produced
by \cref{eq:supp-step} is $z_s = (1-s) z_0 + s\, \epsilon'$ with
$\epsilon' \sim \mathcal{N}(0, I)$ independent of $z_0$, for every
$\nu \in [0, 1]$.
\end{proposition}
\begin{proof}
Substituting, $z_s = (1-s) z_0 + s \big( \sqrt{1-\nu^2}\, \epsilon + \nu\, \xi \big)$.
The two noises are independent standard Gaussians and independent of $z_0$, so
$\epsilon' = \sqrt{1-\nu^2}\, \epsilon + \nu\, \xi$ is Gaussian with mean zero
and covariance $(1 - \nu^2) I + \nu^2 I = I$, and is independent of $z_0$.
\end{proof}
By induction over the steps, a chain run with exact estimates visits, at every
level $s$, a state with the signal coefficient $1 - s$ and the total noise level
$s$ of \cref{eq:supp-interp}, whatever the value of $\nu$; in particular the
final state at $s = 0$ carries no noise. This is the reason for adding the
noise in this way: the network was trained on states of the form
\cref{eq:supp-interp}, and the reward is computed on the decoded final state.
With a learned velocity the estimates are not exact, so the statement concerns
the coefficients of the sampler and not the exact law of the generated states;
the same holds for DDIM~\cite{ddim2021}, of which \cref{eq:supp-step} is the
counterpart for the interpolation \cref{eq:supp-interp}.

\begin{proposition}\textup{(Consistency and deterministic limit.)}
\label{prop:euler}
For any velocity $v_\theta$, (i) $z_t = (1-t)\, \hat z_0 + t\, \hat\epsilon$,
and (ii) the mean is
$\mu_\theta(z_t) = a(s)\, z_t + b(t, s)\, v_\theta$ with
$a(s) = 1 - s + s\sqrt{1-\nu^2}$ and
$b(t, s) = s \sqrt{1-\nu^2}\,(1 - t) - (1 - s)\, t$. For $\nu = 0$ the step is
the Euler step $z_s = z_t + (s - t)\, v_\theta$ of the sampling ODE.
\end{proposition}
\begin{proof}
(i) $(1-t)(z_t - t v_\theta) + t (z_t + (1-t) v_\theta) = z_t$. (ii) Collect the
terms of $(1-s)(z_t - t v_\theta) + s\sqrt{1-\nu^2}\,(z_t + (1-t) v_\theta)$. For
$\nu = 0$, $a = 1$, $b = s(1-t) - (1-s)t = s - t$, and $\sigma_s = 0$.
\end{proof}

\begin{proposition}\textup{(KL term.)}
\label{prop:kl}
For a stochastic step, the KL divergence between the transition of the current
policy and that of the frozen Stage-1 model at the same state is
\begin{equation}
\begin{aligned}
  \mathrm{D}_{\mathrm{KL}}\!\big[ \pi_\theta(\cdot \mid z_t) \,\Vert\,
      \pi_{\mathrm{base}}(\cdot \mid z_t) \big]
  &= \frac{\lVert \mu_\theta - \mu_{\mathrm{base}} \rVert^2}{2 \sigma_s^2} \\
  &= \frac{b(t, s)^2}{2 s^2 \nu^2}\,
     \lVert v_\theta - v_{\mathrm{base}} \rVert^2 .
\end{aligned}
\end{equation}
\end{proposition}
\begin{proof}
Both transitions are Gaussian with covariance $\sigma_s^2 I$, and the KL
divergence between $\mathcal{N}(\mu_1, \sigma^2 I)$ and
$\mathcal{N}(\mu_2, \sigma^2 I)$ is $\lVert \mu_1 - \mu_2 \rVert^2 / (2\sigma^2)$.
By \cref{prop:euler} the two means share the term $a(s) z_t$ and differ by
$b(t, s)\,(v_\theta - v_{\mathrm{base}})$.
\end{proof}
The KL term is thus a penalty on the change of the velocity relative to
Stage~1, weighted over the steps by $b(t, s)^2 / (s \nu)^2$; the second term of
\cref{eq:supp-loss} is this quantity divided by $n d$.

\begin{proposition}\textup{(Likelihood and gradient.)}
\label{prop:pg}
(i) With $\omega_i \equiv 1$, $\ell_\theta = \tfrac{1}{n d} \log
\pi_\theta(z_s \mid z_t, x) + \text{const}$, so
$\rho_\theta = \big( \pi_\theta / \pi_{\theta_{\mathrm{old}}} \big)^{1/(n d)}$.
For general weights, $\ell_\theta$ is, up to a constant and the factor
$1 / (d \sum_i \omega_i)$, the log-density of
$\mathcal{N}\big(\mu_\theta, \mathrm{diag}(\sigma_s^2 / \omega_i)\big)$.
(ii) At $\theta = \theta_{\mathrm{old}}$, where $z_s = \mu_\theta + \sigma_s \xi$,
\begin{equation}
  \nabla_\theta \ell_\theta
  = \frac{b(t, s)}{\sigma_s\, d \sum_i \omega_i}
    \sum_{i=1}^{n} \omega_i\, \xi_i^{\top}\, \frac{\partial v_\theta^{i}}{\partial \theta},
\end{equation}
and the gradient of the first term of \cref{eq:supp-loss} is
$- \hat A_g \nabla_\theta \ell_\theta$.
\end{proposition}
\begin{proof}
(i) $\log \mathcal{N}(z_s; \mu_\theta, \sigma_s^2 I) = - \lVert z_s - \mu_\theta
\rVert^2 / (2 \sigma_s^2) + \text{const}$, and \cref{eq:supp-logp} with
$\omega_i \equiv 1$ is this divided by $n d$. For general weights,
$\sum_i \omega_i \lVert z_s^i - \mu_\theta^i \rVert^2 / (2 \sigma_s^2)$ is the
quadratic form of a Gaussian with variance $\sigma_s^2 / \omega_i$ on token $i$.
(ii) Differentiating \cref{eq:supp-logp} gives
$\nabla_\theta \ell_\theta = \big( \sigma_s^2 d \sum_i \omega_i \big)^{-1}
\sum_i \omega_i\, (z_s^i - \mu_\theta^i)^{\top} \partial \mu_\theta^i / \partial\theta$;
insert $z_s - \mu_\theta = \sigma_s \xi$ and
$\partial \mu_\theta / \partial\theta = b(t, s)\, \partial v_\theta / \partial\theta$
(\cref{prop:euler}). At $\theta_{\mathrm{old}}$ we have $\rho_\theta = 1$, which
is in the interior of the clipping interval, so both arguments of the minimum
equal $\rho_\theta \hat A_g$ in a neighbourhood and the gradient is
$\hat A_g \nabla_\theta \rho_\theta = \hat A_g \nabla_\theta \ell_\theta$.
\end{proof}
Three consequences follow. First, the first gradient step on a batch of
rollouts is a score-function (REINFORCE) update: the fresh noise $\xi$ of each
step, weighted by the advantage, is the direction in which the velocity is
moved. Second, with $\omega_i < 1$ on covered tokens the update is not the
policy gradient of the sampling policy but that of a policy whose covered
tokens are treated as noisier; the effect is that the reward moves the
uncovered tokens ten times more strongly than fully covered ones, while the
unweighted KL term holds all tokens to Stage~1. Third, because
$\rho_\theta$ is the $n d$-th root of the likelihood ratio with
$n d \approx 5 \times 10^5$, a clipping interval of $\pm 0.2$ on $\rho_\theta$ is
never reached in practice (we measured $|\rho_\theta - 1|$ of about $10^{-3}$ or
less); the size of the update is controlled by the KL term, the learning rate
and the number of gradient steps per batch, not by the clip.

\begin{proposition}\textup{(Subsampling the steps.)}
\label{prop:subset}
Let $\mathcal{T}$ be a uniformly random subset of size $m$ of the $M$
stochastic steps, drawn independently of the trajectory. Then
$\mathbb{E}_{\mathcal{T}} \big[ \sum_{k \in \mathcal{T}} g_k \big] =
\tfrac{m}{M} \sum_{k=1}^{M} g_k$ for any per-step gradients $g_k$.
\end{proposition}
\begin{proof}
Each step belongs to $\mathcal{T}$ with probability $m / M$; apply linearity of
expectation.
\end{proof}
Using $7$ of $27$ steps therefore estimates the gradient over all stochastic
steps up to the constant factor $7/27$, at about a quarter of the cost; over
the $4$ gradient steps a given transition is used at least once with probability
$1 - (20/27)^4 \approx 0.70$.

\begin{proposition}\textup{(Both presentation orders.)}
\label{prop:order}
Suppose that, for a pair $(g, h)$, the judge answers on the merits with
probability $1 - \delta$, giving the first image the expected score $q_{gh}$
when $g$ is first and $1 - q_{gh}$ when $h$ is first, and with probability
$\delta$ answers from the position alone: ``first'' with probability $a$,
``second'' with probability $b$, a tie otherwise. Then the expected two-order
score of $g$ against $h$ in \cref{eq:supp-win} is
$(1 - \delta)\, q_{gh} + \delta / 2$.
\end{proposition}
\begin{proof}
On the merits the two orders give $g$ the expected scores $q_{gh}$ and
$1 - (1 - q_{gh}) = q_{gh}$. From the position alone, $g$ receives
$a + \tfrac12 (1 - a - b)$ when shown first and $b + \tfrac12 (1 - a - b)$ when
shown second, which average to $\tfrac12$.
\end{proof}
The expected win rates are thus the same increasing affine function of the
position-free win rates for all candidates, whatever the values of $a$ and $b$;
the ordering of the candidates and the standardized values in
\cref{eq:supp-adv}, which are invariant to such a map, are those of the
position-free win rates. If each pair were asked in one fixed order, candidate
$g$ would instead gain or lose $\delta (a - b)/2$ per comparison according to
the position it happened to be shown in. The model covers a preference that
depends on the position only; it does not cover a preference that interacts
with the content.

\paragraph{What is not guaranteed.} The advantage \cref{eq:supp-adv} subtracts
the mean of a group that contains the candidate itself and divides by the
group's standard deviation, so, as in GRPO, it is not an unbiased estimate of
an advantage. The factor $1 - p_{\mathrm{D}} \in [0, 1]$ is one number per
group: it changes how much each group contributes and leaves the direction of
the update within a group unchanged. Neither the judge nor $p_{\mathrm{D}}$
makes the reward a measure of correctness; \cref{sec:exp-judge} reports how far
the judge agrees with human rankings. Finally, rollouts and test-time sampling
use the same family of steps but not the same setting (uniform levels and
$\nu = 0.5$ against shifted levels and $\nu = 0.9$), so the states on which
the policy is updated are not exactly those visited at test time.

\subsection{Relation to prior work}
\label{sec:supp-rlmath-related}



\paragraph{Policy gradients for image generators.} DDPO~\cite{ddpo2023} treats
the Gaussian steps of a stochastic diffusion sampler as a policy and applies
policy gradients. Flow-matching samplers are deterministic, so
Flow-GRPO~\cite{flowgrpo2025} and DanceGRPO~\cite{dancegrpo2025} first make them
stochastic by replacing the sampling ODE with an SDE that has the same
marginals, and then apply GRPO~\cite{grpo2024}: a group of samples per prompt,
an advantage standardized within the group, and a clipped ratio; Flow-GRPO adds
a KL term to the reference model, in closed form in the velocity, as in
\cref{prop:kl}; DanceGRPO updates a random $60\%$ of the steps and uses no KL
term. \Cref{eq:supp-loss} has this form. Differences in our implementation: the
candidates of a group start from independent noise (DanceGRPO and Pref-GRPO
share the initial noise within a group); we use all $27$ stochastic steps of a
$28$-step rollout and draw $7$ of them per gradient step; and the likelihood is
averaged per dimension, with the consequence for clipping stated after
\cref{prop:pg}.

\paragraph{The sampling step.} In the notation of \cref{eq:supp-split}, the
Euler--Maruyama step of the Flow-GRPO SDE with noise scale $\varsigma_t$ and
step size $\Delta = t - s$ reads, as rewritten by~\cite{cps2025},
$z_s = (1-s)\, \hat z_0 + \big( s - \varsigma_t^2 \Delta / (2t) \big)\, \hat\epsilon
+ \varsigma_t \sqrt{\Delta}\; \xi$. Its total noise level is
$\big[ s^2 + \varsigma_t^2 \Delta^2 / t + (\varsigma_t^2 \Delta / (2t))^2 \big]^{1/2} > s$
for $\varsigma_t > 0$: the step leaves more noise than the level $s$ calls for,
which~\cite{cps2025} identify as the cause of residual noise in the rewarded
images, and for which they propose coefficients-preserving sampling,
$z_s = (1-s)\, \hat z_0 + s \cos(\eta\pi/2)\, \hat\epsilon + s \sin(\eta\pi/2)\, \xi$.
Our step \cref{eq:supp-step} is this update with $\nu = \sin(\eta\pi/2)$
($\nu = 0.5$ corresponds to $\eta = 1/3$), and \cref{prop:cp} is the
coefficients-preserving property in our notation; we make no claim to the
sampler. Two details differ. \cite{cps2025} drop the factor $1 / (2\sigma_s^2)$
from the log-likelihood and from the KL term; we keep it, so that both are the
quantities of \cref{prop:kl,prop:pg} and steps at low noise levels weigh more.
The last step, where $\sigma_s = 0$, is deterministic and excluded in our
implementation.

\paragraph{The reward.} Pref-GRPO~\cite{prefgrpo2025} argues that scores given
to single images differ too little within a group, so that standardization
amplifies noise, and replaces them with the win rate of each image in pairwise
comparisons by a preference reward model, on FLUX.1-dev with the Flow-GRPO
sampler. Our reward \cref{eq:supp-win} has the same form. It differs in what is
compared and how: the judge is a general-purpose vision--language model given a
written rubric, not a model trained for preference; it sees perspective crops
around the region that no conditioning view covers, never the whole image; ties are
allowed and count as half a win; and groups that the judge cannot rank are down-weighted by
the test of \cref{eq:supp-davidson}, where Pref-GRPO weighs all groups equally.
OneReward~\cite{onereward2025} also uses a vision--language model that compares
two images, for mask-guided filling, extension, object removal and text
rendering: a model fine-tuned on preference data compares the current sample
with a sample of a reference model, and the probability of the answer ``yes''
is raised by back-propagating through a one-step prediction of the clean image
from a randomly chosen intermediate step. We use a policy gradient,
which needs neither a differentiable judge nor a reference sample, and we do
not fine-tune the judge. RePainter~\cite{repainter2025} applies GRPO to
object-removal inpainting with a composite reward.

\paragraph{Where the update acts.} Recent image-editing methods also confine a
GRPO update to part of the image: Edit-GRPO~\cite{editgrpo2026} gives the regions inside and
outside the edit mask separate optimization signals, and SpatialFlow-GRPO~\cite{spatialflowgrpo2026}
aligns region-level advantages with the corresponding latent positions, using a
region-aware reward model. In UR-GRPO the region is not an edit mask but the
part of the panorama that no conditioning view covers, the weights $\omega_i$
in \cref{eq:supp-logp} act on its tokens, and the reward is measured by a
prompted judge on crops around the same region; \cref{prop:pg} states what this
weighting does and does not correspond to.

\paragraph{Panorama generation from several views and panoramas for video.}
PanoDiff~\cite{panodiff2023} generates a panorama from one or more unregistered photos: it
predicts their angles, places them on a partial panorama and completes it with
a latent diffusion model conditioned on that panorama and a text prompt.
MultiViewPano~\cite{multiviewpano2025} generates a panorama from one or more arbitrarily positioned
images without training, with a pretrained multi-view diffusion model and
pose-aware stitching. JoPano~\cite{jopano2025} unifies text-to-panorama and view-to-panorama
generation in one DiT on the cubemap through a condition-switching mechanism.
Our model differs in taking any subset of the views of a four-camera rig,
including none, as extra tokens of one fine-tuned FLUX.1-dev, and in its
reinforcement-learning stage. Look Beyond~\cite{lookbeyond2025} generates a panorama from one
image and uses keyframes taken from it as anchors for a video diffusion model,
keeping novel views consistent along long trajectories; we use panoramas in the
same spirit, as shared backgrounds for multi-shot video.

\section{Data, training and test details}
\label{sec:supp-impl}

\paragraph{Sources.} Structured3D~\cite{structured3d} contributes $65{,}474$
renders of indoor rooms in all three furniture settings (empty, simple and
full); Pano360~\cite{spec2021} contributes $30{,}735$ panoramas (mostly real
Flickr photos, plus $714$ synthetic renders); PolyHaven~\cite{polyhaven} and
Humus~\cite{humus} together contribute $981$. We assume the panoramas are
gravity-aligned and do not re-canonicalize them.

\paragraph{Preprocessing.} Each panorama is first resized to
$2048{\times}4096$ (most source images are only $512{\times}1024$ and are
upsampled); conditioning views are rendered from this image by gnomonic
projection and the target is downsampled from it.

\paragraph{Captions.} Qwen3-VL-32B writes each caption from six $90^\circ$
perspective crops of the panorama (four around the horizon, plus up and down).
It is asked for the type of place, the main objects and furniture, the
architectural style and the lighting, and generation stops at $80$ tokens
(median $62$ words). The Laval and SUN360 test panoramas are captioned with the
same crops and token limit.

\paragraph{Stage-1/Stage-2 split.} Structured3D is split by building, not by
image: it has three furniture versions per room, and an image-level split would
let Stage~2 judge rooms already seen in Stage~1. The other sources are
split per image. Stage~1 uses $88{,}062$
images ($90.6\%$ of the data), including $59{,}437$ from $3{,}181$ Structured3D
buildings; Stage~2 uses the remaining $9{,}128$, including $6{,}037$ from the
other $319$ buildings.

\paragraph{Stage-1 training.} Ours and \ourspose are trained identically:
$120$k steps of full fine-tuning (AdamW, learning rate $2{\times}10^{-5}$,
global batch $128$, bf16 compute with fp32 master weights, timestep shift
$t \mapsto st/(1{+}(s{-}1)t)$ with $s{=}3.16$). The caption enters through
FLUX's text encoders (T5 and CLIP) and is replaced by an empty string for
$10\%$ of the samples at every $K$, including $K{=}0$. $20\%$ of the samples
that contain at least one view (one epoch in five for each image, $16\%$ of
all samples) keep the camera layout in the reference orientation, with the
small perturbations of the camera directions.

\paragraph{Camera layout during training.} Training views have a horizontal
field of view between $101^\circ$ and $138^\circ$; each camera direction is
perturbed by up to $\pm 5^\circ$ in latitude and longitude, neighbouring views
all overlap with probability $0.6$, and the layout is rotated uniformly in
$[0^\circ, 360^\circ)$ about the vertical axis and then by up to $\pm 60^\circ$
about the horizontal axis through azimuth $0^\circ$, with each view rendered
without in-plane roll.

\paragraph{Stage-2 training.} The LoRA adapter is trained for $2{,}500$
steps (AdamW, learning rate $10^{-4}$). \Cref{sec:supp-stage2} gives the remaining settings.

\paragraph{Baseline.} Our re-implementation of 360Anything~\cite{anything360}
is trained on FLUX.1-dev for $50$k steps on the same Stage-1 training data as
our models (AdamW, learning rate $5{\times}10^{-5}$, global batch $512$; random
square views with field of view $30^\circ$--$120^\circ$, pitch up to
$\pm 60^\circ$ and roll up to $\pm 15^\circ$; the target is encoded with
circular padding as in the original method). It is trained in a single run
from the FLUX.1-dev weights; the number of steps, batch size and learning rate
are those of~\cite{anything360}, except that our learning rate is constant,
without the $1$k-step warm-up. At test time it takes one
$90^\circ$ square view and decodes with circular padding at the VAE boundary. Compared with the original
paper~\cite{anything360}, our evaluation samples with $28$ steps and text-only
CFG at $3.5$ instead of $50$ steps with CFG $2.0$ on text and $1.5$ on image,
and uses our own preprocessing (for example the HDR tone mapping of Laval).

\paragraph{Test inputs.} The test panorama is resampled bicubically to
$2048{\times}4096$, and the views are rendered from it at $288{\times}512$ with
horizontal field of view $128.1^\circ$ (front), $125.9^\circ$ (back),
$126.3^\circ$ (top) and $123.2^\circ$ (bottom), with no rotation or jitter of
the view directions.

\paragraph{Sampling and guidance.} The test-time sampler uses $28$ steps,
timestep shift $s{=}3.16$ and $\nu{=}0.9$ for all models. The guidance
embedding is always $3.5$, and CFG, where used, has scale $3.5$ with an empty
caption in the unconditional branch.

\paragraph{Test sets.} Laval Indoor~\cite{lavalindoor} is converted from HDR
to LDR at $1024{\times}2048$ by scaling each panorama so that the $99$th
percentile of its luminance maps to $1$, applying a gamma of $1/2.2$ and
clipping to $[0,1]$; it is disjoint from our training data. \emph{Laval
held-out} is the $1{,}889$ Laval scenes that remain after removing the $473$
scenes we used during development. The SUN360 set combines the training and test
parts of the room-layout set ($446 + 72 = 518$ panoramas, available at
$512{\times}1024$; all model outputs are resized to this resolution for
evaluation). To find duplicates, we compared all $31{,}716$
training panoramas that do not come from Structured3D with the test panoramas
(correlation of $64{\times}32$ thumbnails over horizontal rotation, mirroring
and small vertical shifts, and visual inspection of the highest-scoring pairs).
The $2$ SUN360 duplicates are both from Pano360.

\paragraph{Metrics.} Crops have a $90^\circ$ field of view and
$299{\times}299$ pixels, at random azimuths; crop FID and KID use InceptionV3
features and CLIP-FID uses CLIP ViT-B/32 features, and KID is multiplied by
$100$. The range $[110^\circ, 250^\circ]$ keeps crops away from our $K{=}1$
conditioning view when the output is aligned with it. FAED uses the features
of an autoencoder trained on panoramas, on panoramas resized to
$512{\times}1024$. CLIP score uses the average CLIP ViT-B/16 feature of the
$10$ crops, with the caption truncated to CLIP's $77$-token limit. Our DS
implementation differs from earlier papers, so DS is comparable only within
this paper; the ground truth itself has a DS of $0.366$ on Laval and $0.839$ on
SUN360.

\section{Stage-2 implementation details}
\label{sec:supp-stage2}

\paragraph{Sampler.} Rollouts use $28$ uniformly spaced steps of the stochastic
sampler, which adds fresh noise at every step except the last ($\nu{=}0.5$).

\paragraph{Inputs.} Each iteration draws eight inputs. The camera layout
is rotated as in Stage~1, but without reference-orientation samples and with
neighbouring views always overlapping. An input is redrawn if the ground truth
of its uncovered region is too dark or too flat to rank (in its best crop, mean
brightness below $55$ or mean gradient below $0.6$ on a $0$--$255$ scale) or if
the region has no usable crop.

\paragraph{Crops and prompt.} Each input gets two crops with a field of view of
$78.75^\circ$ at $448{\times}448$ pixels; a field of view that followed the size
of the region would change the pixel density by up to $3.6\times$ between
regions of different sizes. Each crop contains a connected part of
$\mathcal{H}$, outlined, together with some of the covered area around it, so
that the judge can see whether the generated content fits its surroundings; we
take no crops of covered regions alone. The prompt contains
the first $300$ characters of the caption. For each crop all $G(G{-}1)/2$ pairs
are asked in both orders; a win gives one point, and a tie or an unparseable
answer gives both candidates half a point.

\paragraph{Policy update.} The test for $p_{\mathrm{D}}$ uses a $\chi^2$
distribution with $G{-}1$ degrees of freedom; it is computed per crop, and the
smallest value over the crops of an input is used. $\mathcal{T}_g$ contains $7$ of the $27$ stochastic
sampler steps, drawn separately for each candidate and for each of the $4$
gradient passes made over one batch of rollouts, and $\beta{=}0.01$. The
log-likelihood is a per-dimension average in which tokens in the uncovered
region have weight $1$ and tokens covered by a conditioning view have weight $0.1$;
the KL term is unweighted. The ratio between the current and the sampling
policy stays within about $10^{-3}$ of one, so the clip never fired.

\section{Full results}
\label{sec:supp-full}

\Cref{tab:full-laval,tab:full-sun} list every input setting, both of our
models before and after Stage~2, and our 360Anything re-implementation at
$K{=}1$, with all metrics. ``Stage 1+2'' is the model after $2{,}500$ Stage-2
steps. At $K{=}0$ without CFG the outputs
look hazy; we list these rows for completeness.

\begin{table*}[t]
\centering\footnotesize
\setlength{\tabcolsep}{4pt}
\begin{tabular}{@{}ll ccccccc@{}}
\toprule
Input & Model & crop-FID$\downarrow$ & KID$\downarrow$ & CLIP-FID$\downarrow$ & FAED$\downarrow$ & CS$\uparrow$ & DS & raw-ERP FID$\downarrow$ \\
\midrule
\multirow{4}{*}{$K{=}0$} & Ours, Stage 1 & $86.03$ & $5.69$ & $17.84$ & $29.50$ & $26.18$ & $0.783$ & $87.11$ \\
 & Ours, Stage 1+2 & $\mathbf{33.27}$ & $\mathbf{1.90}$ & $\mathbf{10.00}$ & $16.20$ & $\mathbf{30.25}$ & $0.996$ & $\mathbf{41.58}$ \\
 & \ourspose, Stage 1 & $91.31$ & $6.26$ & $18.60$ & $17.06$ & $26.51$ & $0.825$ & $93.50$ \\
 & \ourspose, Stage 1+2 & $41.38$ & $2.55$ & $11.31$ & $\mathbf{15.69}$ & $29.59$ & $0.867$ & $52.93$ \\
\midrule
\multirow{4}{*}{$K{=}0$ + CFG} & Ours, Stage 1 & $24.33$ & $1.19$ & $\mathbf{7.49}$ & $16.94$ & $30.19$ & $0.840$ & $42.80$ \\
 & Ours, Stage 1+2 & $\mathbf{16.57}$ & $\mathbf{0.62}$ & $8.15$ & $\mathbf{16.13}$ & $31.39$ & $0.967$ & $\mathbf{34.39}$ \\
 & \ourspose, Stage 1 & $26.61$ & $1.45$ & $9.67$ & $20.75$ & $31.63$ & $0.941$ & $41.14$ \\
 & \ourspose, Stage 1+2 & $19.89$ & $0.91$ & $9.37$ & $20.32$ & $\mathbf{32.46}$ & $0.969$ & $34.95$ \\
\midrule
\multirow{5}{*}{$K{=}1$} & 360Anything (re-impl.) & $30.66$ & $1.92$ & $8.16$ & $6.25$ & $28.38$ & $0.310$ & $42.95$ \\
 & Ours, Stage 1 & $34.25$ & $2.16$ & $10.85$ & $5.71$ & $27.74$ & $0.565$ & $33.01$ \\
 & Ours, Stage 1+2 & $\mathbf{13.44}$ & $\mathbf{0.64}$ & $\mathbf{3.78}$ & $\mathbf{4.71}$ & $\mathbf{30.17}$ & $0.695$ & $\mathbf{19.73}$ \\
 & \ourspose, Stage 1 & $45.41$ & $3.06$ & $13.12$ & $6.12$ & $27.53$ & $0.554$ & $31.61$ \\
 & \ourspose, Stage 1+2 & $19.47$ & $1.07$ & $5.94$ & $5.55$ & $29.78$ & $0.600$ & $21.24$ \\
\midrule
\multirow{5}{*}{$K{=}1$ + CFG} & 360Anything (re-impl.) & $11.59$ & $0.50$ & $2.66$ & $8.38$ & $30.76$ & $0.365$ & $27.28$ \\
 & Ours, Stage 1 & $14.91$ & $0.68$ & $3.62$ & $7.91$ & $30.28$ & $0.731$ & $25.81$ \\
 & Ours, Stage 1+2 & $\mathbf{10.20}$ & $\mathbf{0.35}$ & $\mathbf{2.43}$ & $6.62$ & $31.51$ & $0.866$ & $21.76$ \\
 & \ourspose, Stage 1 & $17.76$ & $0.91$ & $3.45$ & $4.83$ & $31.05$ & $0.837$ & $23.02$ \\
 & \ourspose, Stage 1+2 & $14.09$ & $0.64$ & $3.13$ & $\mathbf{4.80}$ & $\mathbf{32.04}$ & $0.921$ & $\mathbf{19.76}$ \\
\midrule
\multirow{4}{*}{$K{=}2$} & Ours, Stage 1 & $4.88$ & $0.16$ & $1.65$ & $4.20$ & $30.11$ & $0.655$ & $21.50$ \\
 & Ours, Stage 1+2 & $3.62$ & $\mathbf{0.07}$ & $\mathbf{1.03}$ & $\mathbf{3.77}$ & $30.57$ & $0.680$ & $16.93$ \\
 & \ourspose, Stage 1 & $4.16$ & $0.13$ & $1.45$ & $4.04$ & $30.34$ & $0.700$ & $19.77$ \\
 & \ourspose, Stage 1+2 & $\mathbf{3.57}$ & $0.09$ & $1.12$ & $3.89$ & $\mathbf{30.60}$ & $0.683$ & $\mathbf{16.84}$ \\
\midrule
\multirow{4}{*}{$K{=}3$} & Ours, Stage 1 & $3.81$ & $0.11$ & $1.22$ & $3.98$ & $30.38$ & $0.672$ & $16.20$ \\
 & Ours, Stage 1+2 & $\mathbf{3.13}$ & $\mathbf{0.06}$ & $\mathbf{0.89}$ & $\mathbf{3.60}$ & $\mathbf{30.68}$ & $0.691$ & $14.89$ \\
 & \ourspose, Stage 1 & $3.74$ & $0.11$ & $1.25$ & $3.85$ & $30.42$ & $0.706$ & $15.43$ \\
 & \ourspose, Stage 1+2 & $3.28$ & $0.08$ & $1.03$ & $3.73$ & $30.63$ & $0.688$ & $\mathbf{14.45}$ \\
\midrule
\multirow{4}{*}{$K{=}4$} & Ours, Stage 1 & $3.17$ & $0.08$ & $1.05$ & $0.19$ & $30.42$ & $0.667$ & $7.49$ \\
 & Ours, Stage 1+2 & $\mathbf{2.80}$ & $\mathbf{0.05}$ & $\mathbf{0.86}$ & $\mathbf{0.13}$ & $\mathbf{30.64}$ & $0.670$ & $7.13$ \\
 & \ourspose, Stage 1 & $3.19$ & $0.08$ & $1.07$ & $0.15$ & $30.43$ & $0.689$ & $6.38$ \\
 & \ourspose, Stage 1+2 & $2.94$ & $0.06$ & $0.92$ & $\mathbf{0.13}$ & $30.61$ & $0.687$ & $\mathbf{6.13}$ \\
\bottomrule
\end{tabular}
\caption{Laval Indoor, all inputs, models and metrics. Crops at azimuths $90^\circ$--$270^\circ$. KID is multiplied by $100$. Bold: best within each input setting, except DS (ground-truth DS is $0.366$, so a lower value is not necessarily better). Raw-ERP FID is computed directly on the equirectangular image and is given only as a cross-check.}
\label{tab:full-laval}
\end{table*}

\begin{table*}[t]
\centering\footnotesize
\setlength{\tabcolsep}{4pt}
\begin{tabular}{@{}ll ccccccc@{}}
\toprule
Input & Model & crop-FID$\downarrow$ & KID$\downarrow$ & CLIP-FID$\downarrow$ & FAED$\downarrow$ & CS$\uparrow$ & DS & raw-ERP FID$\downarrow$ \\
\midrule
\multirow{4}{*}{$K{=}0$} & Ours, Stage 1 & $92.10$ & $8.11$ & $36.97$ & $23.78$ & $27.23$ & $0.717$ & $125.99$ \\
 & Ours, Stage 1+2 & $\mathbf{49.18}$ & $\mathbf{4.21}$ & $\mathbf{17.54}$ & $\mathbf{20.30}$ & $\mathbf{29.30}$ & $0.921$ & $\mathbf{63.52}$ \\
 & \ourspose, Stage 1 & $87.50$ & $8.00$ & $32.60$ & $28.28$ & $27.52$ & $0.804$ & $126.09$ \\
 & \ourspose, Stage 1+2 & $50.26$ & $4.48$ & $18.67$ & $24.98$ & $29.16$ & $0.847$ & $75.28$ \\
\midrule
\multirow{4}{*}{$K{=}0$ + CFG} & Ours, Stage 1 & $40.41$ & $3.08$ & $16.10$ & $\mathbf{14.50}$ & $29.12$ & $0.762$ & $82.38$ \\
 & Ours, Stage 1+2 & $\mathbf{35.63}$ & $\mathbf{2.63}$ & $\mathbf{11.54}$ & $14.80$ & $29.61$ & $0.876$ & $\mathbf{63.79}$ \\
 & \ourspose, Stage 1 & $46.63$ & $4.15$ & $15.51$ & $39.56$ & $29.80$ & $0.926$ & $79.41$ \\
 & \ourspose, Stage 1+2 & $41.09$ & $3.45$ & $12.87$ & $36.86$ & $\mathbf{30.06}$ & $0.938$ & $65.42$ \\
\midrule
\multirow{5}{*}{$K{=}1$} & 360Anything (re-impl.) & $34.06$ & $1.69$ & $10.97$ & $8.07$ & $27.98$ & $0.475$ & $68.59$ \\
 & Ours, Stage 1 & $43.31$ & $2.70$ & $16.12$ & $7.04$ & $27.27$ & $0.670$ & $63.03$ \\
 & Ours, Stage 1+2 & $\mathbf{18.45}$ & $\mathbf{0.58}$ & $\mathbf{6.49}$ & $5.30$ & $28.25$ & $0.796$ & $\mathbf{43.87}$ \\
 & \ourspose, Stage 1 & $42.91$ & $2.65$ & $15.03$ & $5.78$ & $27.52$ & $0.668$ & $59.79$ \\
 & \ourspose, Stage 1+2 & $20.70$ & $0.79$ & $7.55$ & $\mathbf{4.46}$ & $\mathbf{28.39}$ & $0.748$ & $46.94$ \\
\midrule
\multirow{5}{*}{$K{=}1$ + CFG} & 360Anything (re-impl.) & $\mathbf{18.35}$ & $\mathbf{0.69}$ & $\mathbf{5.73}$ & $5.58$ & $28.70$ & $0.529$ & $60.20$ \\
 & Ours, Stage 1 & $23.03$ & $1.06$ & $7.71$ & $4.38$ & $28.21$ & $0.791$ & $55.64$ \\
 & Ours, Stage 1+2 & $21.11$ & $0.87$ & $5.87$ & $\mathbf{3.42}$ & $28.47$ & $0.941$ & $46.40$ \\
 & \ourspose, Stage 1 & $23.17$ & $1.08$ & $6.75$ & $4.38$ & $28.85$ & $0.911$ & $49.44$ \\
 & \ourspose, Stage 1+2 & $22.71$ & $1.08$ & $6.55$ & $4.27$ & $\mathbf{28.98}$ & $0.987$ & $\mathbf{45.63}$ \\
\midrule
\multirow{4}{*}{$K{=}2$} & Ours, Stage 1 & $9.97$ & $0.32$ & $3.69$ & $2.31$ & $27.71$ & $0.846$ & $45.94$ \\
 & Ours, Stage 1+2 & $9.07$ & $\mathbf{0.24}$ & $\mathbf{3.15}$ & $2.01$ & $27.81$ & $0.868$ & $\mathbf{38.55}$ \\
 & \ourspose, Stage 1 & $9.46$ & $0.28$ & $3.48$ & $1.88$ & $27.78$ & $0.904$ & $42.41$ \\
 & \ourspose, Stage 1+2 & $\mathbf{9.01}$ & $\mathbf{0.24}$ & $3.22$ & $\mathbf{1.62}$ & $\mathbf{27.87}$ & $0.907$ & $38.90$ \\
\midrule
\multirow{4}{*}{$K{=}3$} & Ours, Stage 1 & $9.27$ & $0.28$ & $3.25$ & $1.92$ & $27.70$ & $0.839$ & $38.24$ \\
 & Ours, Stage 1+2 & $8.69$ & $\mathbf{0.22}$ & $\mathbf{2.86}$ & $1.82$ & $27.83$ & $0.864$ & $35.01$ \\
 & \ourspose, Stage 1 & $9.06$ & $0.27$ & $3.16$ & $1.48$ & $27.75$ & $0.943$ & $35.23$ \\
 & \ourspose, Stage 1+2 & $\mathbf{8.64}$ & $0.23$ & $2.97$ & $\mathbf{1.29}$ & $\mathbf{27.85}$ & $0.944$ & $\mathbf{33.55}$ \\
\midrule
\multirow{4}{*}{$K{=}4$} & Ours, Stage 1 & $7.91$ & $0.22$ & $2.89$ & $0.85$ & $27.71$ & $0.889$ & $20.87$ \\
 & Ours, Stage 1+2 & $7.83$ & $0.21$ & $\mathbf{2.68}$ & $0.84$ & $\mathbf{27.79}$ & $0.887$ & $20.75$ \\
 & \ourspose, Stage 1 & $7.80$ & $0.20$ & $2.83$ & $0.53$ & $27.72$ & $1.028$ & $17.68$ \\
 & \ourspose, Stage 1+2 & $\mathbf{7.67}$ & $\mathbf{0.19}$ & $2.70$ & $\mathbf{0.45}$ & $27.77$ & $1.012$ & $\mathbf{17.60}$ \\
\bottomrule
\end{tabular}
\caption{SUN360 indoor subset, all inputs, models and metrics (internal comparison only). Crops at azimuths $90^\circ$--$270^\circ$. KID is multiplied by $100$. Bold: best within each input setting, except DS (ground-truth DS is $0.839$, so a lower value is not necessarily better). Raw-ERP FID is computed directly on the equirectangular image and is given only as a cross-check.}
\label{tab:full-sun}
\end{table*}

\paragraph{Crop ranges and held-out scenes.} \Cref{tab:k1crop} repeats the
crop-FID comparison at $K{=}1$ with crops kept away from our conditioning view
and on the held-out Laval scenes (\cref{sec:exp-main}).

\begin{table}[t]
\centering\footnotesize
\setlength{\tabcolsep}{2.5pt}
\resizebox{\linewidth}{!}{%
\begin{tabular}{@{}l cc cc cc@{}}
\toprule
& \multicolumn{2}{c}{Laval (all)} & \multicolumn{2}{c}{Laval (held-out)} & \multicolumn{2}{c}{SUN360} \\
\cmidrule(lr){2-3}\cmidrule(lr){4-5}\cmidrule(lr){6-7}
Method & wide & away & wide & away & wide & away \\
\midrule
\multicolumn{7}{@{}l}{\textit{Guidance embedding only}} \\
360Anything (re-impl.) & $30.66$ & $31.82$ & $30.79$ & $31.93$ & $34.06$ & $36.11$ \\
Ours, Stage 1 & $34.25$ & $37.98$ & $34.78$ & $38.51$ & $43.31$ & $48.50$ \\
Ours, Stage 1+2 & $\mathbf{13.44}$ & $\mathbf{14.67}$ & $\mathbf{14.24}$ & $\mathbf{15.55}$ & $\mathbf{18.45}$ & $\mathbf{20.29}$ \\
\ourspose, Stage 1 & $45.41$ & $52.75$ & $45.95$ & $53.29$ & $42.91$ & $48.60$ \\
\ourspose, Stage 1+2 & $19.47$ & $21.78$ & $20.20$ & $22.59$ & $20.70$ & $22.53$ \\
\midrule
\multicolumn{7}{@{}l}{\textit{With CFG (scale $3.5$)}} \\
360Anything (re-impl.) & $11.59$ & $12.40$ & $12.26$ & $13.27$ & $\mathbf{18.35}$ & $\mathbf{19.86}$ \\
Ours, Stage 1 & $14.91$ & $16.08$ & $15.60$ & $16.99$ & $23.03$ & $25.14$ \\
Ours, Stage 1+2 & $\mathbf{10.20}$ & $\mathbf{10.90}$ & $\mathbf{10.86}$ & $\mathbf{11.77}$ & $21.11$ & $22.91$ \\
\ourspose, Stage 1 & $17.76$ & $19.61$ & $18.41$ & $20.33$ & $23.17$ & $25.11$ \\
\ourspose, Stage 1+2 & $14.09$ & $15.33$ & $14.79$ & $16.22$ & $22.71$ & $24.50$ \\
\bottomrule
\end{tabular}}
\caption{Crop FID at $K{=}1$ under two crop ranges and on held-out scenes. \emph{wide}: crops at azimuths $90^\circ$--$270^\circ$, as in~\cite{anything360}; \emph{away}: azimuths $110^\circ$--$250^\circ$, so that no crop overlaps our conditioning view when the output is aligned with it (\cref{sec:exp-setup}). \emph{Laval (held-out)} excludes the $473$ Laval scenes used during development. Bold: best in each group.}
\label{tab:k1crop}
\end{table}

\paragraph{Published numbers.} \Cref{tab:reported} lists the numbers reported
in~\cite{anything360} for reference.

\begin{table}[t]
\centering\footnotesize
\setlength{\tabcolsep}{4pt}
\begin{tabular}{@{}l ccccc@{}}
\toprule
Method & FID$\downarrow$ & KID$\downarrow$ & CLIP-FID$\downarrow$ & FAED$\downarrow$ & CS$\uparrow$ \\
\midrule
OmniDreamer~\cite{omnidreamer2022} & $71.0$ & $5.17$ & $23.9$ & $19.2$ & -- \\
PanoDiffusion~\cite{panodiffusion2024} & $58.6$ & $4.08$ & $26.6$ & $106.8$ & -- \\
Diffusion360~\cite{diffusion360} & $33.1$ & $2.07$ & $16.9$ & $23.7$ & $26.38$ \\
CubeDiff~\cite{cubediff2025} & $9.5$ & $0.32$ & $3.2$ & $18.4$ & $27.02$ \\
360Anything~\cite{anything360} & $8.0$ & $0.22$ & $4.6$ & $9.8$ & $29.21$ \\
\bottomrule
\end{tabular}
\caption{Published numbers on Laval Indoor at $K{=}1$, copied from Table~1 of~\cite{anything360} (which in turn copies the first four rows from~\cite{cubediff2025}). They come from different evaluation code and preprocessing and are listed for reference only; they should not be compared directly with \cref{tab:main}.}
\label{tab:reported}
\end{table}

\section{More examples across view counts}
\label{sec:supp-examples}

\begin{figure*}[t]
\centering
\begin{minipage}[t]{0.495\textwidth}\centering
{\small\textbf{Indoor}}\\[2pt]
\setlength{\tabcolsep}{1pt}\renewcommand{\arraystretch}{0.5}
\begin{tabular}{cc}
{\scriptsize Ground truth} & {\scriptsize $K{=}0$ + CFG}\\
\includegraphics[width=0.49\linewidth]{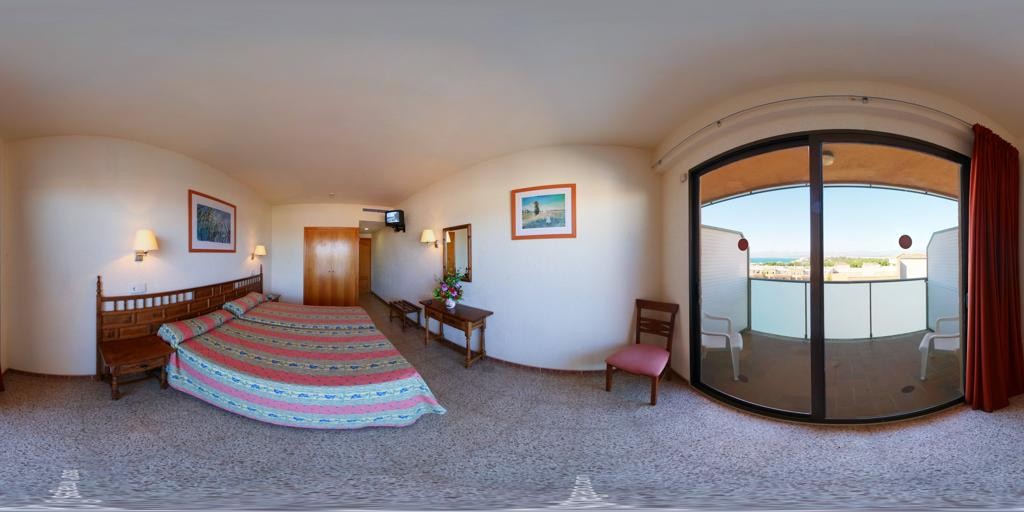} &
\includegraphics[width=0.49\linewidth]{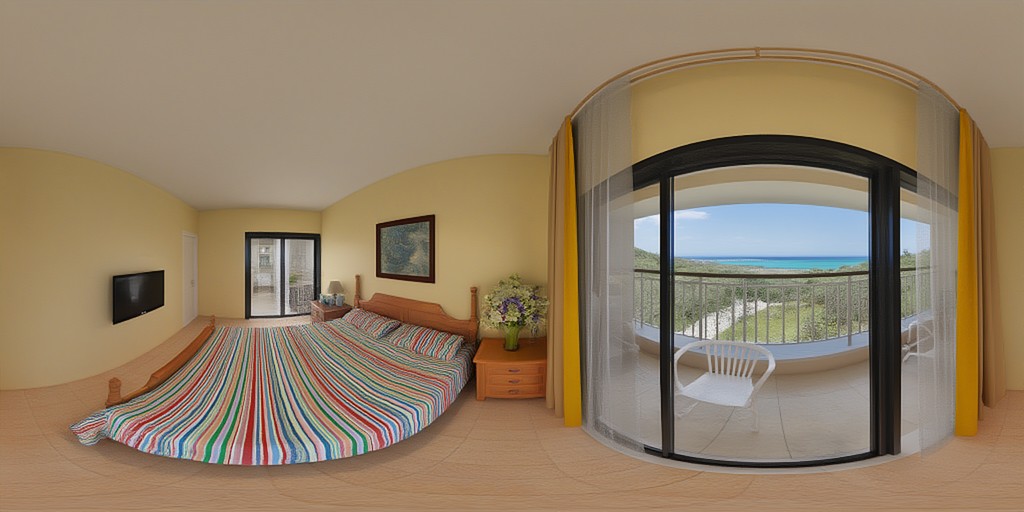}\\[3pt]
{\scriptsize $K{=}1$ + CFG} & {\scriptsize $K{=}2$}\\
\includegraphics[width=0.49\linewidth]{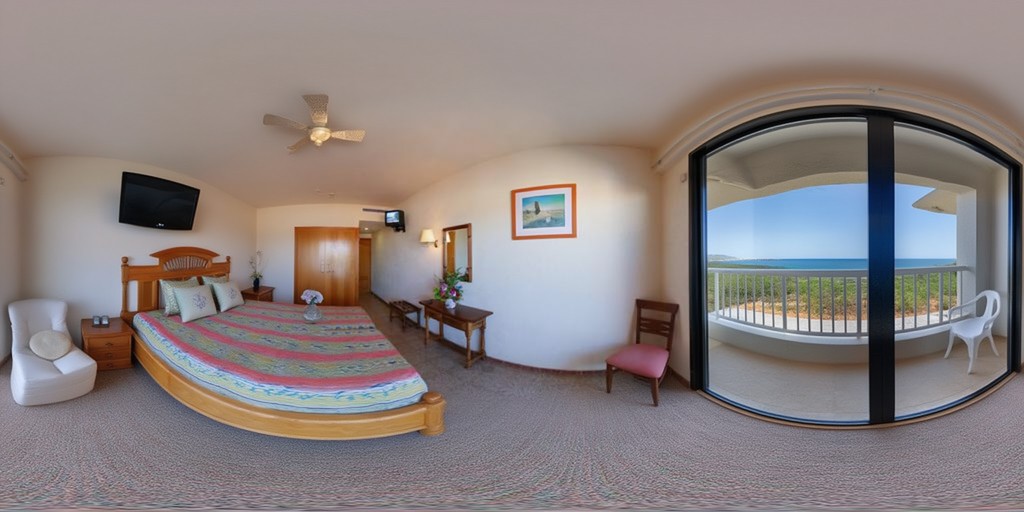} &
\includegraphics[width=0.49\linewidth]{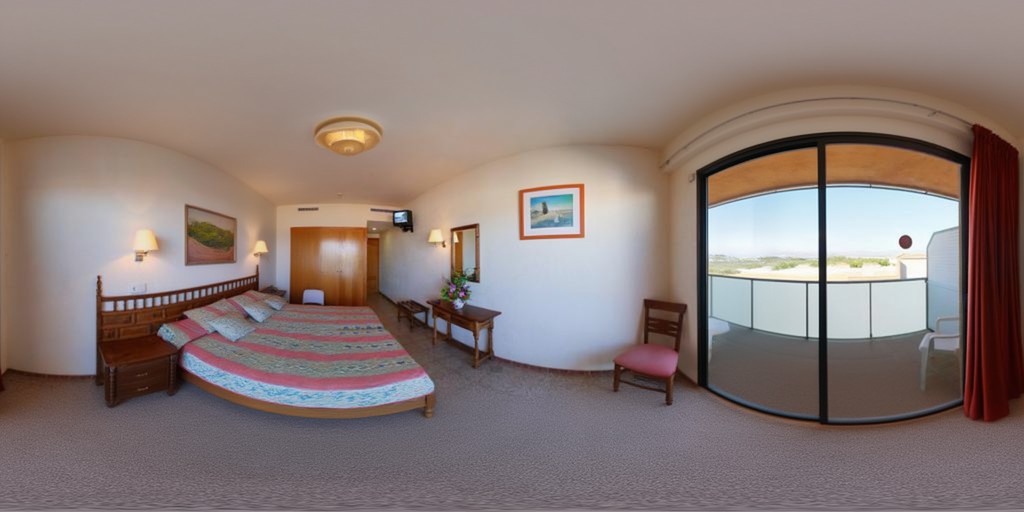}\\[3pt]
{\scriptsize $K{=}3$} & {\scriptsize $K{=}4$}\\
\includegraphics[width=0.49\linewidth]{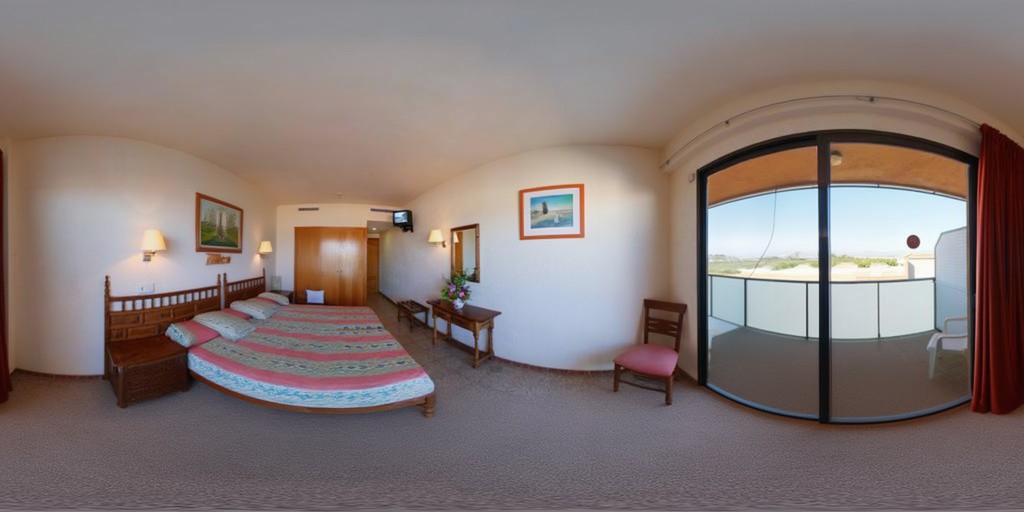} &
\includegraphics[width=0.49\linewidth]{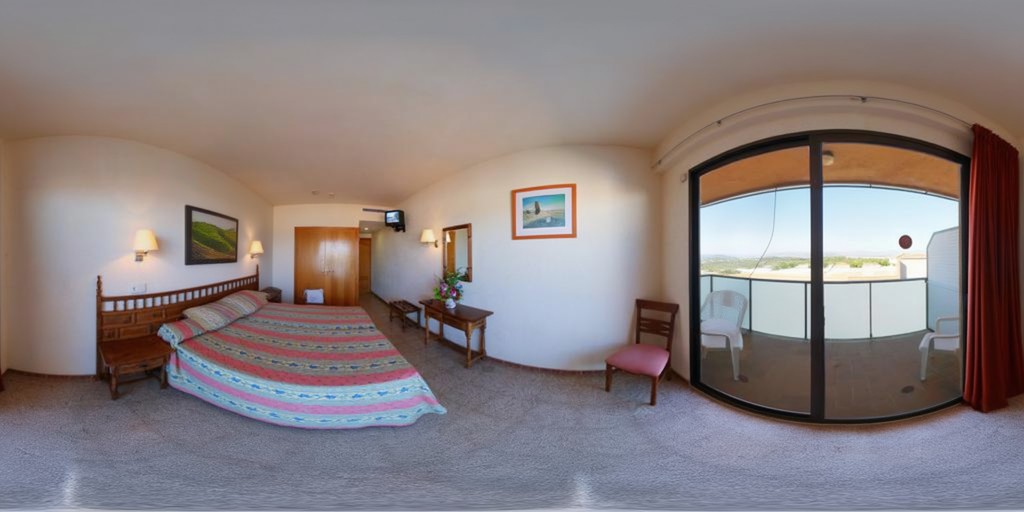}\\[3pt]
\end{tabular}
\end{minipage}\hfill
\begin{minipage}[t]{0.495\textwidth}\centering
{\small\textbf{Outdoor}}\\[2pt]
\setlength{\tabcolsep}{1pt}\renewcommand{\arraystretch}{0.5}
\begin{tabular}{cc}
{\scriptsize Ground truth} & {\scriptsize $K{=}0$ + CFG}\\
\includegraphics[width=0.49\linewidth]{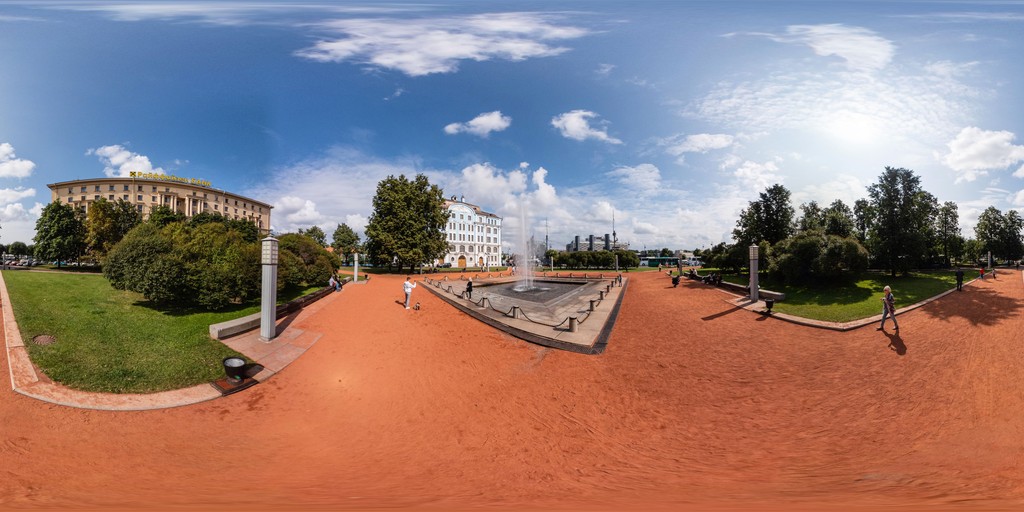} &
\includegraphics[width=0.49\linewidth]{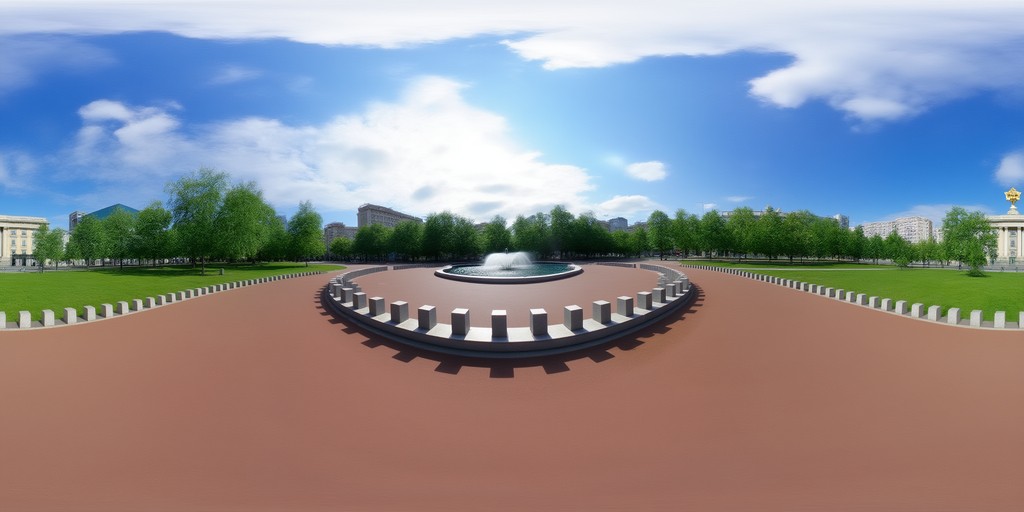}\\[3pt]
{\scriptsize $K{=}1$ + CFG} & {\scriptsize $K{=}2$}\\
\includegraphics[width=0.49\linewidth]{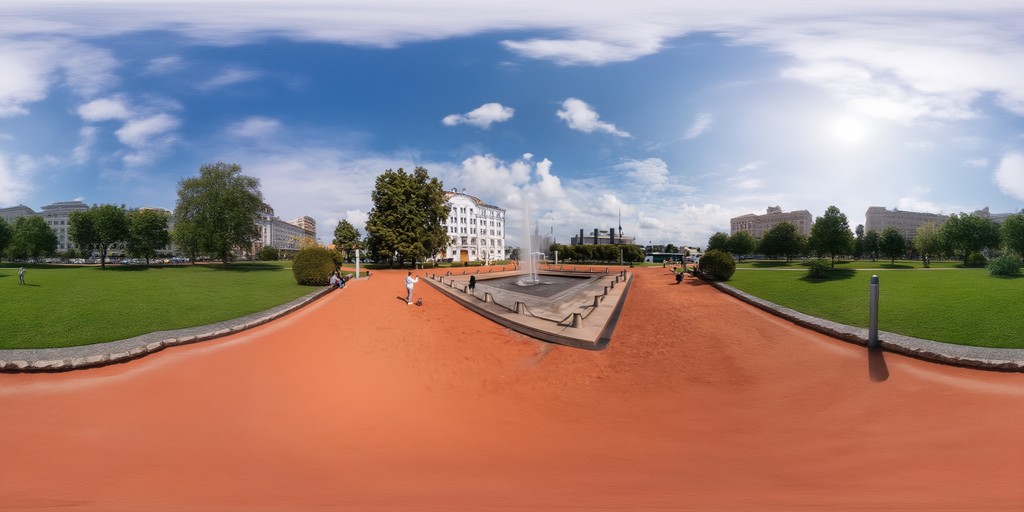} &
\includegraphics[width=0.49\linewidth]{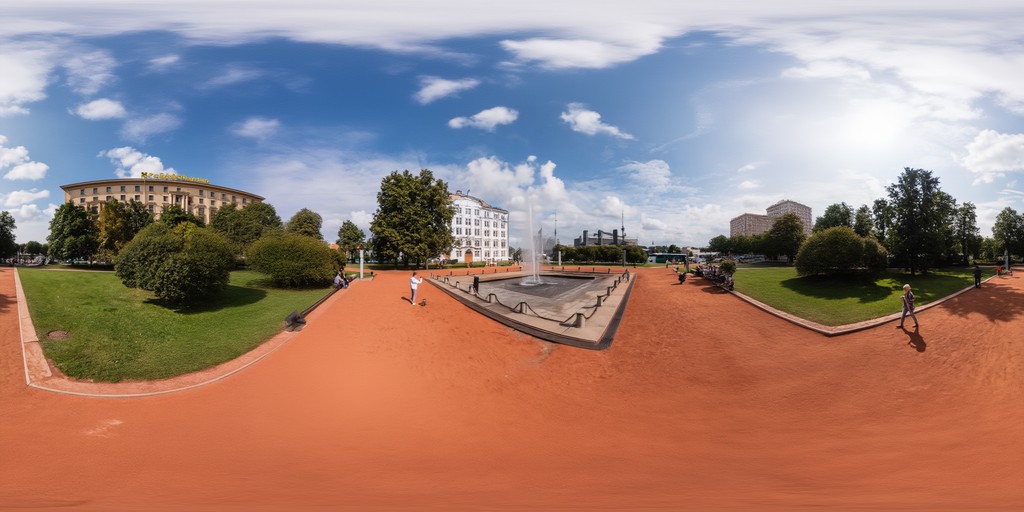}\\[3pt]
{\scriptsize $K{=}3$} & {\scriptsize $K{=}4$}\\
\includegraphics[width=0.49\linewidth]{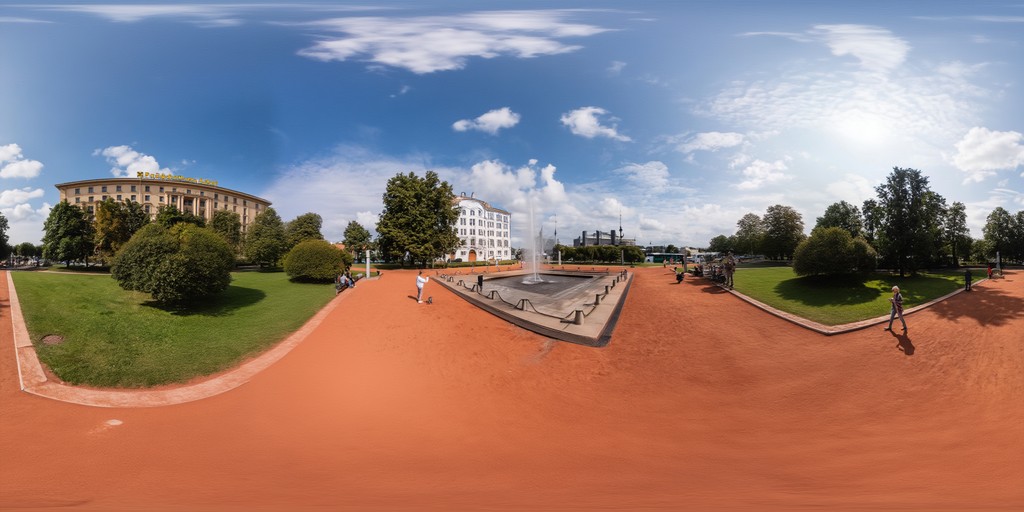} &
\includegraphics[width=0.49\linewidth]{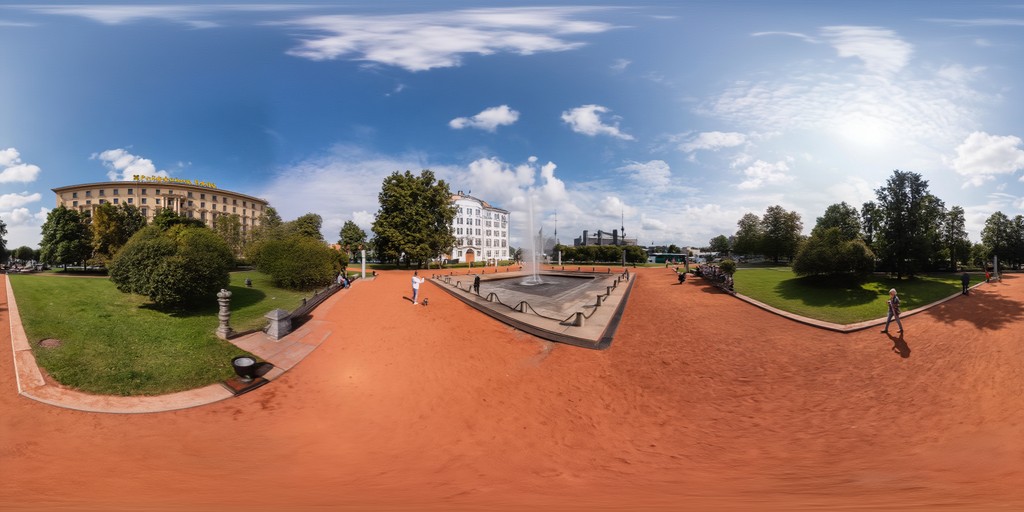}\\[3pt]
\end{tabular}
\end{minipage}
\caption{Ours (Stage~1+2) at every number of conditioning views $K$, with the
same caption and noise seed within each panorama. Neither panorama is used in
training; the indoor one is from the SUN360 test set. Left: ``a cozy, well-lit
bedroom with \dots\ a private balcony \dots\ a scenic view of the sea''. Right:
``a sunny, open urban park \dots\ a central square with a modern fountain''
(captions shortened). At $K{=}0$ the model sees only the caption, so its
panorama need not resemble the ground truth.}
\label{fig:supp-examples}
\end{figure*}

\Cref{fig:supp-examples} shows Ours (Stage~1+2) on two more panoramas at every
number of conditioning views, one indoor and one outdoor. The conditioning
views are those of
\cref{sec:exp-setup} (front; front and back; front, back and top; all four),
and $K{=}0$ and $K{=}1$ use CFG. The outdoor example is qualitative only; both
of our quantitative test sets contain indoor scenes only.

\section{Judge validation details}
\label{sec:supp-judge}

\paragraph{Differences from Stage~2.} The candidates in both batches come from
an earlier fixed-layout Stage-1 model, all at $K{=}4$, with crops whose field of
view follows the size of the uncovered region (as in Stage~2, they include some
of the covered area around it). The prompt is a shorter validation wording, not
the one used for the Stage-2 reward, and \cref{tab:judge} keeps only
order-consistent verdicts, while the reward uses win rates over both orders.

\paragraph{Human labels.} The two batches are $20$ groups of eight candidates
from the deterministic sampler and $40$ groups of eight from the stochastic
sampler ($\nu{=}0.9$), with regions filtered for brightness and content.
Rankings allow ties and are never aggregated across groups. In the second pass
the annotator reviewed and revised their scores, changing $28$ of $317$ cells;
the two passes agree on $512$ of $517$ pairs, a self-consistency of $0.990$
$[0.978, 0.996]$.

\paragraph{\Cref{tab:judge}.} The $40$ groups give $1{,}120$ pairs, each asked
in both orders; $70$ to $312$ pairs remain after keeping only order-consistent
verdicts on pairs the human did not tie. The pairs/group column also counts
pairs the human tied.

\paragraph{Position preference.} With the discouraging tie wording, Qwen's
win rate for the candidate shown first is $0.667$; with the neutral wording it
is $0.277$. InternVL prefers the second position under both wordings ($0.426$
and $0.200$). Order-consistent filtering removes such verdicts, since a judge
that only prefers one position can never be order-consistent; Stage~2 instead
asks every pair in both orders, so position preference cancels in the win rate,
a combination we did not validate against the human labels.

\paragraph{Gemma4-12B.} Gemma4-12B is excluded not because it is weak but
because it disagrees with humans: random guessing would give $0.5$, and its
order-consistent verdicts fall below that, so adding it would lower agreement.

\section{Orientation measurement}
\label{sec:supp-orient}

We estimate each output's offset about the vertical axis by sliding the
ground-truth content around the front view over the output (zero-normalized
cross-correlation of $256{\times}128$ grayscale images, band
$|\text{latitude}| < 45^\circ$, $338$ Laval scenes, $270$ of them held-out). At $K{=}1$ with CFG, Ours
after Stage~2 is within $10^\circ$ in $48\%$ of scenes. Stage~2 changes the
fractions reported in \cref{sec:exp-pose} by at most two points.

\section{Figure details}
\label{sec:supp-figsel}

\paragraph{\Cref{fig:qual}.} The conditioning view of our models is a $16{:}9$ image with a horizontal field
of view of $128^\circ$ at the panorama centre; crops have a $90^\circ$ field of
view, and azimuths are measured from the panorama centre, increasing to the
right. The seam strip is cut from the panorama rolled by $180^\circ$. The Ours
outputs shown are within about $7^\circ$ of the ground-truth orientation; the
re-implementation places its input $77^\circ$ and $110^\circ$ to the right of
the centre, so in the second row its $135^\circ$ crop, and in the first row the
left part of that crop, show its input region.
The scenes come from the held-out scenes on which orientation was measured
whose Ours outputs (both stages) are within $10^\circ$ of the ground-truth
orientation with a correlation peak of at least $0.5$ (\cref{sec:exp-pose};
$111$ of $270$, of which we leave out four shown in
\cref{fig:spectrum,fig:rl}): they are the three with the brightest ground truth
(mean luminance within $60^\circ$ of the horizon) of different room types, of
which the figure shows two for space, the
room type being the first room word in the caption (two ice rinks, whose
captions name no room, are skipped). This is one of five disjoint sets of three
scenes, each chosen by a different fixed rule that favours bright ground truth
and uses only the ground truth, its caption and orientation estimates, never a
comparison of the outputs; we picked this set after viewing all five.

The black band at the bottom of the ground truth in \cref{fig:qual} is part of
the Laval images.

\paragraph{\Cref{fig:spectrum}.} At $K{=}2$ the red lines enclose the two
covered regions (the back view's region wraps around the left and right edges);
at $K{=}3$ the covered part lies above the line; at $K{=}4$ they enclose the two
remaining uncovered side regions. The scene is one of two picked after viewing $15$
candidates taken at a fixed stride from the $173$ scenes, out of the $270$
held-out scenes on which orientation was measured, whose generated panoramas at
$K{=}2$, $3$ and $4$ all lie within $10^\circ$ of the ground-truth
orientation; $176$ of the $270$ pass at $K{=}2$ ($256$ at $K{=}3$, $266$ at
$K{=}4$), so the $K{=}2$ row shows an aligned output only.

\paragraph{\Cref{fig:rl}.} We picked these two rows from $42$ scene--$K$
pairs ($41$ held-out scenes listed by a fixed rule, restricted to scenes whose
output orientation was measured to be within $10^\circ$).
The pairs are at $K{=}2$ and $K{=}3$ only ($21$ each); this was fixed before
the candidates were viewed, and no $K{=}1$ or $K{=}4$ crop was viewed when the
rows were chosen. At $K{=}1$ the conditioning view covers $24\%$ of the sphere,
so the same crop is $77\%$ uncovered, with the boundary near one edge of the
crop; the crops of Ours in \cref{fig:qual} lie entirely in the uncovered region
and compare the two stages at $K{=}1$ (with CFG), and Stage~2 lowers crop FID
far more at this setting than at $K{=}2$--$4$
(\cref{tab:main,tab:spectrum}). At $K{=}4$ the uncovered region is two patches
of about $53^\circ{\times}55^\circ$ ($13\%$ of the sphere together), and
Stage~2 changes crop FID least (\cref{tab:spectrum}).

\paragraph{\Cref{fig:pose}.} We show the first two scenes of a fixed list of
$48$ where both statements of the caption hold, which they do in $98\%$ and
$100\%$ of the list.

\section{Panoramas as shared backgrounds for shot images}
\label{sec:supp-storyboard}

\begin{figure*}[t]
\centering
\begin{minipage}[b]{0.66\textwidth}\centering
\includegraphics[width=\linewidth]{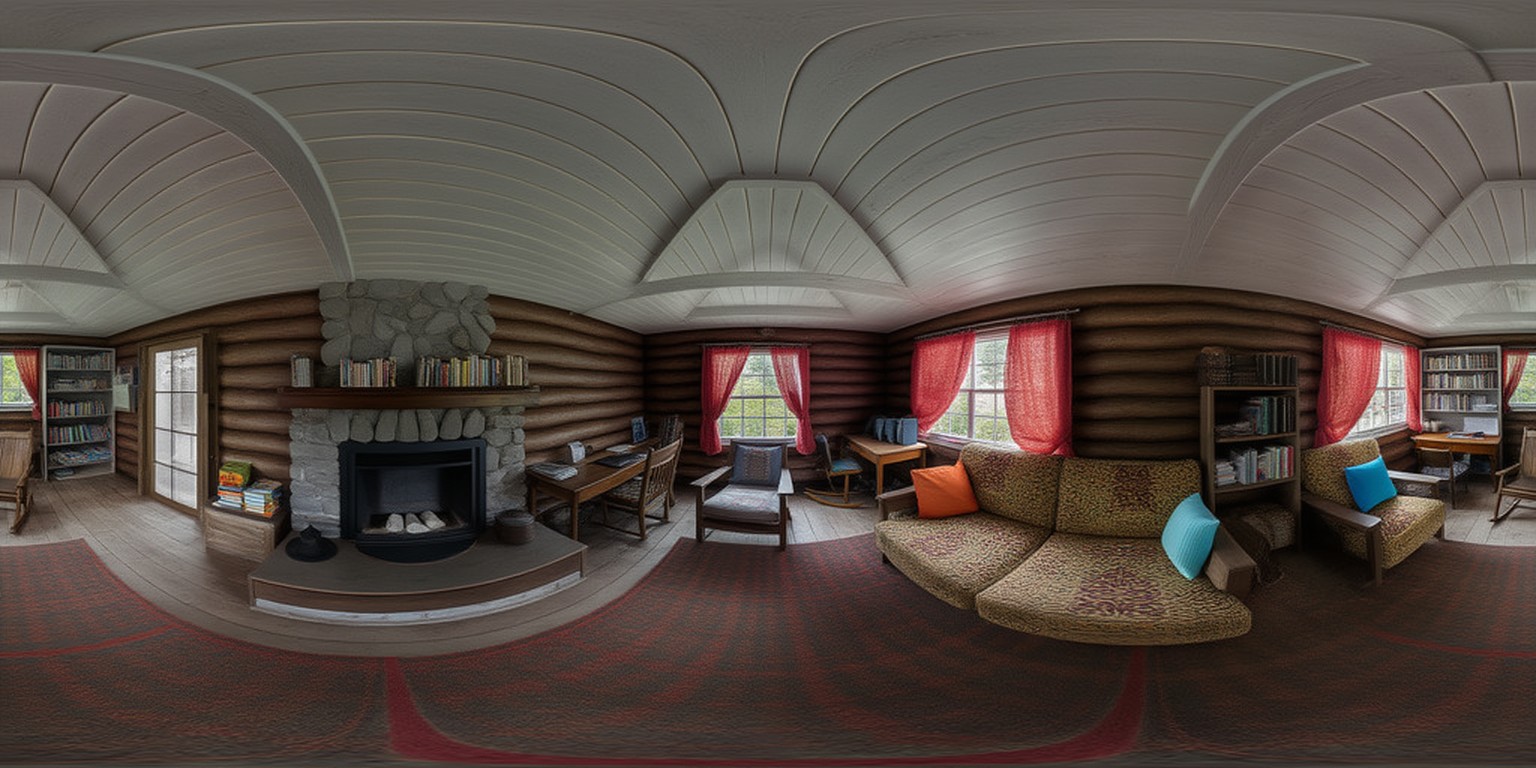}\\[-1pt]
{\scriptsize Panorama from Ours ($K{=}0$, caption only, with CFG)}
\end{minipage}\hfill
\begin{minipage}[b]{0.32\textwidth}\centering
\includegraphics[width=\linewidth]{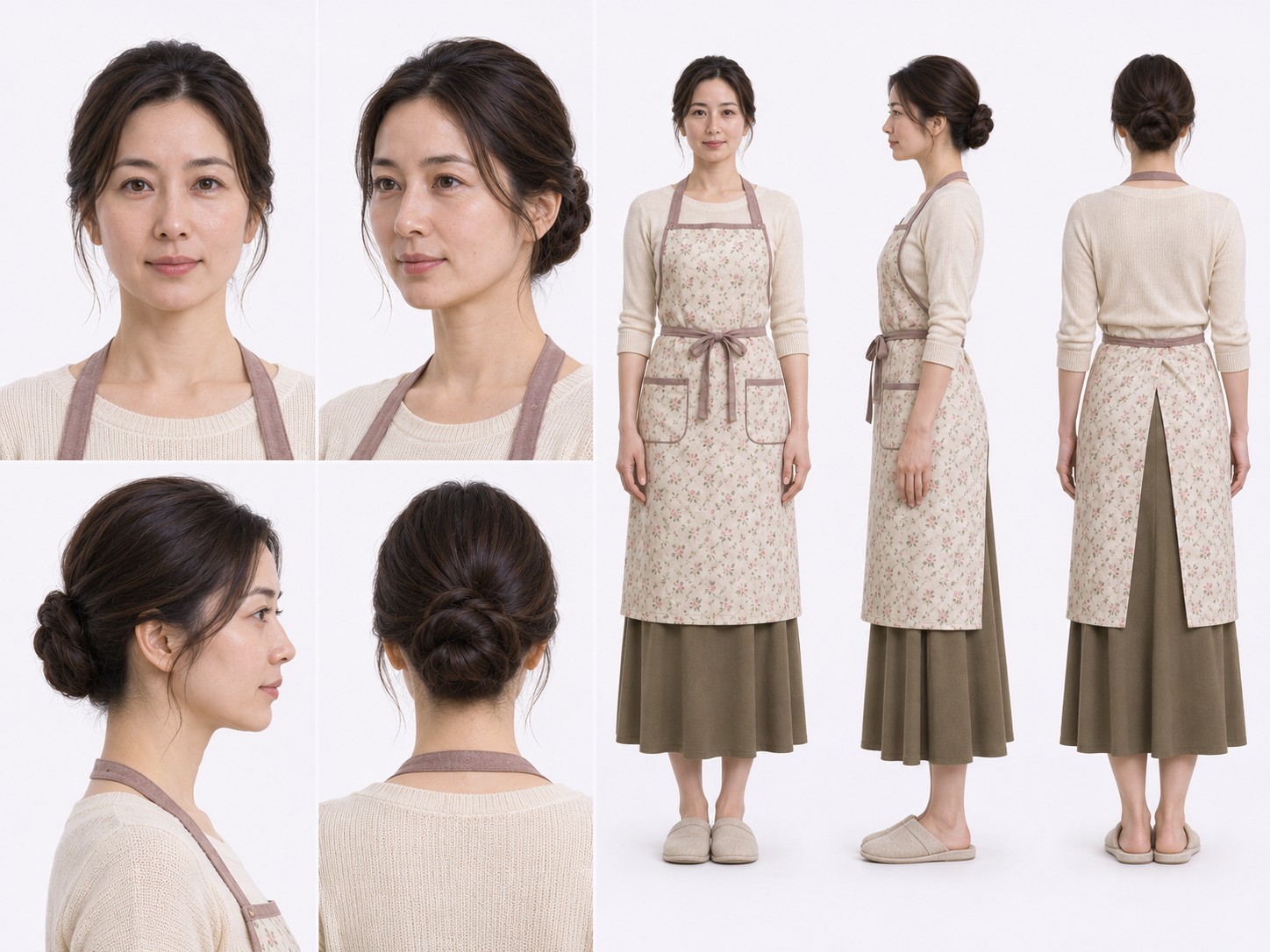}\\[-1pt]
{\scriptsize Views of the character}
\end{minipage}\\[4pt]
\begin{minipage}[t]{0.245\textwidth}\centering
\includegraphics[width=\linewidth]{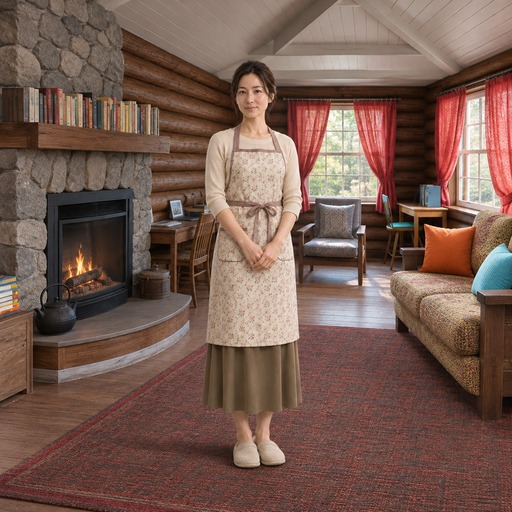}\\[-1pt]{\scriptsize Shot 1: wide view of the living room}
\end{minipage}\hfill
\begin{minipage}[t]{0.245\textwidth}\centering
\includegraphics[width=\linewidth]{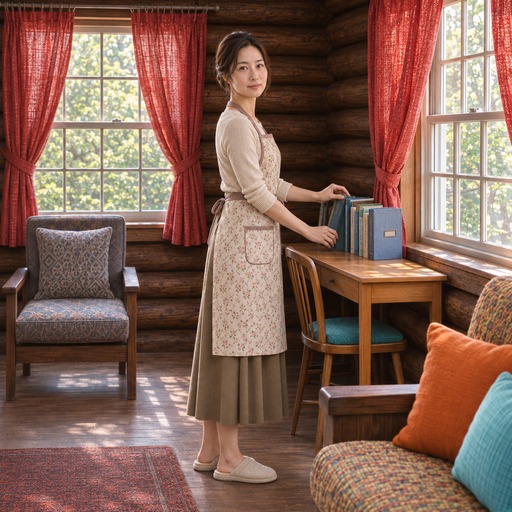}\\[-1pt]{\scriptsize Shot 2: tidying books by the window}
\end{minipage}\hfill
\begin{minipage}[t]{0.245\textwidth}\centering
\includegraphics[width=\linewidth]{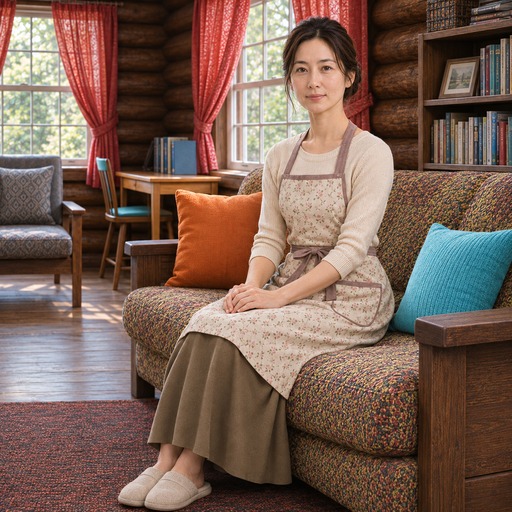}\\[-1pt]{\scriptsize Shot 3: sitting on the sofa}
\end{minipage}\hfill
\begin{minipage}[t]{0.245\textwidth}\centering
\includegraphics[width=\linewidth]{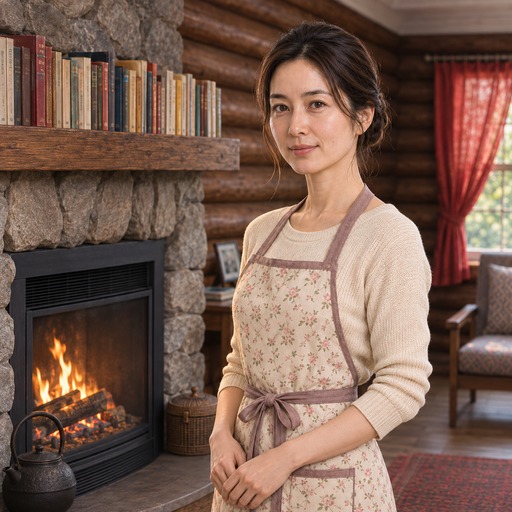}\\[-1pt]{\scriptsize Shot 4: close-up by the fireplace}
\end{minipage}
\caption{One panorama from our model, generated from a caption alone ($K{=}0$,
with CFG; ``a cozy, rustic interior with exposed wooden log walls and a
white-beamed ceiling \dots''), used as the shared background of four shot
images. The views of the character and the shot images ($512{\times}512$) are
generated with GPT-image~\cite{openaiimage}; each shot image is generated from
the views of the character, the panorama and a text prompt, and the four shots
show the same room from different camera angles.}
\label{fig:storyboard}
\end{figure*}

\Cref{fig:storyboard} shows the use that motivates this work
(\cref{sec:intro}): one panorama from our model serves as the background of
the images that start several shots of the same scene, each seen from a
different camera angle, so the setting stays the same from shot to shot. Each
shot image is generated with GPT-image~\cite{openaiimage} from views of the
character, the panorama and a text prompt, and a video generation model then produces the video from these images,
with a cut at each of them.

\end{document}